\documentclass{article}

\usepackage{microtype}
\usepackage{graphicx}
\usepackage{subfigure}
\usepackage{booktabs} 

\usepackage{hyperref}

\usepackage[accepted]{icml2024}

\usepackage{amsmath}
\usepackage{amssymb}
\usepackage{mathtools}
\usepackage{amsthm}

\usepackage{rotating, multirow}

\usepackage[capitalize,noabbrev]{cleveref}

\theoremstyle{plain}

\theoremstyle{definition}

\theoremstyle{remark}

\usepackage{xcolor}
\newcommand{\vect}[1]{{\boldsymbol{#1}}}
\newcommand{\E}{\mathbb{E} }
\newcommand{\R}{\mathbb{R} }
\newcommand{\KL}{D_{\textrm{KL}}}
\newcommand{\gray}[1]{\textcolor{lightgray}{#1}}

\usepackage[textsize=tiny]{todonotes}

\icmltitlerunning{Discovering Multiple Solutions from a Single Task in Offline RL}

\begin{document}

\twocolumn[
\icmltitle{Discovering Multiple Solutions from a Single Task \\ in Offline Reinforcement Learning}




\begin{icmlauthorlist}
\icmlauthor{Takayuki Osa}{ut,riken}
\icmlauthor{Tatsuya Harada}{ut,riken}
\end{icmlauthorlist}

\icmlaffiliation{ut}{The University of Tokyo, Japan}
\icmlaffiliation{riken}{RIKEN Center for Advanced Intelligence Project}

\icmlcorrespondingauthor{Takayuki Osa}{osa@mi.t.u-tokyo.ac.jp}

\icmlkeywords{Machine Learning, ICML}

\vskip 0.3in
]



\printAffiliationsAndNotice{}  

\begin{abstract}
Recent studies on online reinforcement learning (RL) have demonstrated the advantages of learning multiple behaviors from a single task, as in the case of few-shot adaptation to a new environment. 
Although this approach is expected to yield similar benefits in offline RL, appropriate methods for learning multiple solutions have not been fully investigated in previous studies. 
In this study, we therefore addressed the problem of finding multiple solutions from a single task in offline RL.
We propose algorithms that can learn multiple solutions in offline RL, and empirically investigate their performance. Our experimental results show that the proposed algorithm learns multiple qualitatively and quantitatively distinctive solutions in offline RL.
\end{abstract}

\section{Introduction}
The benefits of discovering diverse solutions have been demonstrated in literature pertaining to online reinforcement learning (RL), as in the cases of few-shot adaptation to changes in the environment~\cite{Kumar19} and composition of complex motion by sequencing different behavioral styles~\cite{Kumar20}.
For example, a locomotion task may encompass multiple motion styles using different postures.
Although the discovery of diverse solutions in offline RL is expected to be equally beneficial in practice, it has not been fully investigated in previous studies. Accordingly, we address the problem of discovering diverse solutions in offline RL.

The primary challenge of learning multiple solutions in offline RL is the problem of learning the latent skill space.
In the case of online RL, diverse behaviors are typically modeled with a latent-conditioned policy wherein the value of the latent variable is stored alongside the state action pairs during the data collection process. 
Although offline RL employs a dataset containing the state-action pairs, this dataset does not contain information that indicates the skill type, such as the value of the latent variable. 
Therefore, it is essential to learn the latent skill representations in an unsupervised manner.

\begin{figure}[tb]
	\centering
	\includegraphics[width=0.95\columnwidth]{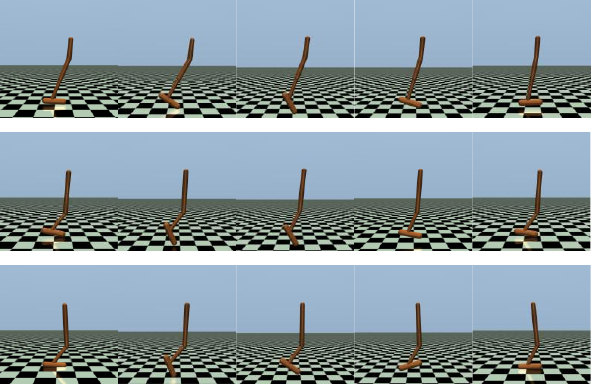}
	\vspace{-0.3cm}
	\caption{Sequential snapshots of multiple locomotion behaviors obtained by proposed method for the hopper agent. }
	\label{fig:AWAC_V_I_diverse_expert_hopper_intro}
	\vspace{-0.4cm}
\end{figure}

In this study, we introduce a novel algorithm designed to uncover multiple solutions through the unsupervised learning of latent skill representations. 
We developed our algorithm to tackle the challenge of identifying multiple solutions in offline RL through a coordinate ascent approach, resembling the expectation-maximization (EM) algorithm.
In our algorithm, we iteratively update a policy conditioned on latent variables to capture multiple behaviors, along with a posterior distribution that models latent skill representations.
To evaluate our algorithm, we constructed datasets that contain diverse behaviors using the D4RL framework~\cite{Fu20}, with locomotion tasks based on MuJoCo~\cite{Todorov12}. 
The experimental results demonstrate that the proposed algorithm discovers multiple solutions as shown in Figure~\ref{fig:AWAC_V_I_diverse_expert_hopper_intro}, and that few-shot adaptation to new environments can be performed by learning multiple behaviors in offline RL.

\section{Background}
\subsection{Related Work}
In literature pertaining RL, it is known that multiple optimal policies may elicit the optimal value function~\cite{Sutton2018}.
The discovery of diverse solutions has been actively investigated in the evolutionary computation community, with such algorithms often referred to as quality-diversity (QD) algorithms~\cite{Chatzilygeroudis21,Cully18,Parker-Holder20,Pugh16}.
These QD algorithms can be adapted to offline RL settings by learning dynamics models as proposed in \cite{Hein18,Hein18b}.
In the context of online RL, studies such as \cite{Kumar20a,Osa22,Sharma20} reported that multiple solutions can be obtained from a single task, 
enabling few-shot adaptation to new environments and composition of complex behavior by sequencing learned behaviors. 
Recent studies on robotics have also reported that multiple solutions frequently manifest in robotics problems~\cite{Osa22a,Toussaint18}.

The task of learning latent representations of the state-action space is closely related to that of discovering multiple solutions in RL. 
Recent studies pertaining to offline RL have investigated the leveraging of latent representations of the state-action space~\cite{Chen22,Zhou20,Osa23}, 
revealing this task to improve the stability of the learning process. 
The study by \cite{Ajay21} proposed a framework to learn primitive skills from offline data and utilize them for downstream tasks.
Although the focus of the study presented by \cite{Ajay21} is utilizing the primitive skills for downstream tasks, their work is relevant to ours in a sense that how to extract multiple behaviors in offline RL is investigated.
\citet{Liu23} recently proposed a framework to leverage expert demonstrations in offline RL.
While the framework proposed by \citet{Liu23} is not specifically designed for finding multiple behaviors, they also demonstrated that multiple behaviors can be found by their method in a unsupervised manner.

Our algorithm utilizes coordinate ascent, drawing inspiration from the EM algorithm~\cite{Bishop06}. 
While the policy update in the E-step can be perceived as an extension of MPO~\cite{Abdolmaleki18} or AWAC~\cite{Nair20}, 
it is essential to note that our algorithm distinguishes itself by the alternate updating of the latent-conditioned policy and posterior distribution. 
This distinctive approach represents a novel algorithm designed specifically for learning multiple solutions in offline RL, setting it apart from existing methodologies.

A concurrent work by \citet{Mao24} recently addressed the problem of finding multiple solutions in offline RL and proposed a method called Stylized Offline RL (SORL).
\citet{Mao24} formulated the problem of finding multiple solutions in offline RL as a clustering problem, 
where the latent variable that characterizes the behavior style is discrete, and the number of solutions obtained by SORL is finite. 
In contrast, our formulation is more general in the sense that the latent variable can be either continuous or discrete. While we report results based on the continuous latent variable in this study, the discrete latent variable can also be learned using techniques such as the Gumbel-softmax trick~\cite{Jang17,Maddison17}.

\subsection{Reinforcement Learning}
In RL, we consider a Markov decision process~(MDP) that consists of a tuple~$(\mathcal{S}, \mathcal{A}, \mathcal{P}, r  , \gamma, d)$ 
where $\mathcal{S}$ is the state space,  $\mathcal{A}$ is the action space,  $\mathcal{P}(\vect{s}_{t+1}|\vect{s}_t, \vect{a}_t)$ is the transition probability density,  
$r(\vect{s}, \vect{a})$ is the reward function, $\gamma$ is the discount factor, and $d(\vect{s}_0)$ is the probability density of the initial state.
A policy $\pi(\vect{a}|\vect{s}): \mathcal{S} \times \mathcal{A} \mapsto \R$ is defined as the conditional probability density over actions given the states.
The aim of RL is to learn a policy that maximizes the expected return $\E[R_0|\pi] $ where $R_t = \sum_{k=t}^{T} \gamma^{k-t}r(\vect{s}_k, \vect{a}_k)$.

In offline RL, it is assumed that the learning agent is provided with a fixed dataset, $\mathcal{D}=\{ (\vect{s}_i, \vect{a}_i, r_i, \vect{s}'_i) \}^N_{i=1}$, comprising states, actions, rewards, and next states collected by an unknown behavior policy $\beta(\vect{a}|\vect{s})$.
For convenience, we denote by $d^\beta(\vect{s})$ the state distribution obtained by executing the behavior policy $\beta$.
The objective of offline RL is to obtain a policy that maximizes the expected return using $\mathcal{D}$ without online interactions with the environment during the learning process.

\subsection{Latent-Conditioned Policy}
A latent-conditioned policy $\pi(\vect{a}|\vect{s}, \vect{z})$ is typically used to model multiple behaviors with a single model, as demonstrated by prior studies on online RL~\cite{Kumar20a,Sharma20}.
Here, the latent variable $\vect{z}$, which may be continuous or discrete, is used to control the behavior type of the latent-conditioned policy  $\pi(\vect{a}|\vect{s}, \vect{z})$.
We assumed that the value of the latent variable is sampled at the beginning of an episode and fixed until the episode's end. 
When the dataset contains multiple behaviors, we assume that the behavior policy can be written as $\beta(\vect{a}|\vect{s}) = \int \beta(\vect{a}|\vect{s}, \vect{z})p(\vect{z})d\vect{z}$, where $p(\vect{z})$ is the prior distribution of the latent variable, and the value of $\vect{z}$ controls the behavior style of the behavior policy.
As the value of $\vect{z}$ is unknown, the primary challenge of learning multiple behaviors in offline RL is to estimate the latent representations.

In the context of online RL, a latent-conditioned action-value function is typically estimated to evaluate the quality of different behaviors.
The latent-conditioned action-value function is defined as
\begin{align}
	Q^\pi(\vect{s}, \vect{a}, \vect{z})= \E_{\pi} \left[ R | \vect{s}, \vect{a}, \vect{z} \right],
\end{align}
which models the expected return when taking action $\vect{a}$ under state $\vect{s}$ with the latent variable $\vect{z}$, and then following policy $\pi$.
The value of the latent variable $\vect{z}$ corresponds to a type of the behavior encoded in the latent-conditioned policy $\pi$.
Similarly, a latent-conditioned state-value function is defined as
\begin{equation}
	V^\pi(\vect{s}, \vect{z})= \E_{\pi} \left[ R | \vect{s}, \vect{z} \right],
\end{equation}
which models the expected return when starting from state $\vect{s}$ with the latent variable $\vect{z}$, and then following policy $\pi$.
In our work, we consider the advantage function given by $A^\pi(\vect{s}, \vect{a}, \vect{z}) = Q^\pi(\vect{s}, \vect{a}, \vect{z}) - V^\pi(\vect{s}, \vect{z})$.

\section{Algorithm to Learn Diverse Behaviors in Offline RL}
When a latent-conditioned policy is trained in an online RL manner, the value of the latent variable is known during the rollout, being stored alongside states, actions, and rewards. 
However, in the context of offline RL, we assume that the value of the latent variable corresponding to a state-action pair is unknown. 
To train a latent-conditioned policy, it is therefore essential to estimate the value of the latent variable in an unsupervised learning manner. 
In addition, as the performance of the latent-conditioned policy relies on the learned latent representations, the encoder that estimates the latent variable corresponding to a state-action pair must reflect the behavior of the policy. 
Therefore, it is essential to jointly train the latent-conditioned policy and encoder.

In this section, we describe our method for obtaining latent representations of the state action space, and explain how to train a latent-conditioned policy using the learned representations.

\subsection{Problem Formulation}
To acquire multiple solutions, we train a latent-conditioned policy $\pi_{\vect{\theta}}(\vect{a}|\vect{s}, \vect{z})$ parametrized by vector $\vect{\theta}$. 
Concurrently, we train a posterior distribution $q_{\vect{\phi}}(\vect{z}|\vect{s}, \vect{a})$ parametrized by vector $\vect{\phi}$ to learn latent skill representations. 
Our objective in solving a given task is to optimize $\pi_{\vect{\theta}}(\vect{a}|\vect{s}, \vect{z})$ and $q_{\vect{\phi}}(\vect{z}|\vect{s}, \vect{a})$ to maximize the expected return.
In addition, we also maximize the mutual information between the latent variable and the state-action pair, $I_{\pi}(\vect{z};\vect{s}, \vect{a})$, to enhance the diversity of the behavior encoded in the policy.
Moreover, to mitigate the generation of out-of-distribution actions in offline RL, we impose a constraint based on the Kullback-Leibler (KL) divergence between the policy $\pi_{\vect{\theta}}$ and the behavior policy $\beta$. 
Consequently, we formulate the problem of discovering multiple solutions in offline RL as the following optimization problem:

\begin{align}
	&\max_{\pi, q}  \left( \E_{\vect{s}\sim d^{\beta}, \vect{a}\sim \pi_{\vect{\theta}}, \vect{z}\sim q_{\vect{\phi}}} \left[  A^{\pi}(\vect{s}, \vect{a}, \vect{z})  \right] + \lambda I_{\pi}(\vect{z};\vect{s}, \vect{a}) \right) \label{eq:problem_obj} \\
	& \textrm{s.t.} \ \E_{\vect{s}\sim d^{\beta}, \vect{z}\sim q_{\vect{\phi}}} \left[ \KL\left( \pi_{\vect{\theta}}(\cdot|\vect{s}, \vect{z})||\beta(\cdot | \vect{s}, \vect{z}) \right) \right] \leq \epsilon_\pi,
	\label{eq:problem}
\end{align}
where $ \lambda$ is a coefficient to balance the weight on each term.
The proposed algorithm addresses maximization of the first term in \eqref{eq:problem_obj} through a procedure resembling the EM algorithm.
In the E-step, we update the latent-conditioned policy $\pi_{\vect{\theta}}(\vect{a}|\vect{s}, \vect{z})$ based on the posterior distribution $q_{\vect{\phi}}(\vect{z}|\vect{s}, \vect{a})$.
In the M-step, we update the posterior distribution $q_{\vect{\phi}}(\vect{z}|\vect{s}, \vect{a})$ given the latent-conditioned policy $\pi_{\vect{\theta}}(\vect{a}|\vect{s}, \vect{z})$.
Subsequently, both $\pi_{\vect{\theta}}(\vect{a}|\vect{s}, \vect{z})$ and $q_{\vect{\phi}}(\vect{z}|\vect{s}, \vect{a})$ are updated to maximize the variational lower bound of the mutual information $I_{\pi}(\vect{z};\vect{s}, \vect{a})$.
Our algorithm can be viewed as a form of coordinate ascent, wherein the latent-conditioned policy $\pi_{\vect{\theta}}(\vect{a}|\vect{s}, \vect{z})$ and the posterior distribution $q_{\vect{\phi}}(\vect{z}|\vect{s}, \vect{a})$ are alternately updated.
In the following section, we provide a detailed description of each step.

\subsection{E-Step}
Given the encoder $q_{\vect{\phi}}(\vect{z} | \vect{s}, \vect{a})$, we solve the following optimization problem in E-step:
\begin{align}
	&\max_{\pi} \E_{\vect{s}\sim d^{\beta}, \vect{a}\sim \pi_{\vect{\theta}}, \vect{z}\sim q_{\vect{\phi}}} \left[  A^{\pi}(\vect{s}, \vect{a}, \vect{z})  \right] \\
	& \ \textrm{s.t.} \ \E_{\vect{s}\sim d^{\beta}, \vect{z}\sim q_{\vect{\phi}}} \left[ \KL\left( \pi_{\vect{\theta}}(\cdot|\vect{s}, \vect{z})||\beta(\cdot | \vect{s}, \vect{z}) \right) \right] \leq \epsilon_\pi.  \label{eq:KL_RL_prob_latent}
\end{align}
The KL divergence constraint in \eqref{eq:KL_RL_prob_latent} encourages the policy close to the data distribution, enabling us to avoid generating out-of-distribution actions.
In addition, when a dataset contains diverse behaviors, the constraint in the above problem maintains the diversity of the learned policy $\pi$.

The solution to this problem is given in a nonparametric form as
\begin{align}
	\pi^*(\vect{a}|\vect{s}, \vect{z}) = \frac{1}{Z_\pi} \beta(\vect{a}|\vect{s}, \vect{z}) \exp \left( \frac{1}{\alpha_\pi}  A^{\pi}(\vect{s}, \vect{a}, \vect{z}) \right),
	\label{eq:E-step-nonpara}
\end{align}
where $Z_\pi$ is the partition function.
To approximate the nonparametric solution in \eqref{eq:E-step-nonpara} with a parameterized model,
we minimize the KL divergence between $\pi^*(\vect{a}|\vect{s}, \vect{z})$ in \eqref{eq:E-step-nonpara} and $\pi_{\vect{\theta}}(\vect{a}|\vect{s}, \vect{z})$:
\begin{align}
	\E_{\vect{s},\vect{z}} \left[\KL\left( \pi^*(\vect{a}|\vect{s}, \vect{z}) || \pi_{\vect{\theta}}(\vect{a}|\vect{s}, \vect{z}) \right) \right].
\end{align}
The minimizer of the above KL divergence can be obtained by maximizing the weighted log-likelihood:
\begin{align}
	&\mathcal{L}_{\textrm{E-step}}(\vect{\theta}) \nonumber \\
	&= \E_{\vect{s},\vect{a}\sim d^\beta, \beta} \E_{\vect{z}\sim q}\left[ 	W_\pi(\vect{s}, \vect{a}, \vect{z}) \log \pi_{\vect{\theta}}(\vect{a}|\vect{s}, \vect{z})  \right],
	\label{eq:awr_latent}
\end{align}
where the weight $W_\pi(\vect{s}, \vect{a}, \vect{z})$ is given by
\begin{align}
	W_\pi(\vect{s}, \vect{a}, \vect{z}) =  \exp \left(\frac{1}{\alpha_\pi} A^\pi(\vect{s}, \vect{a}, \vect{z}) \right).
	\label{eq:weight_pi}
\end{align}
The derivation of \eqref{eq:E-step-nonpara} and \eqref{eq:awr_latent} is presented in Appendix~\ref{app:derivation}. 
According to the derived results, the latent-conditioned policy is updated using \eqref{eq:awr_latent} in the E-step. 
The policy update in \eqref{eq:awr_latent} can be understood as an extension of MPO~\cite{Abdolmaleki18} or AWAC~\cite{Nair20}.

\subsection{M-Step}
In the M-step, we introduce a constraint on the update of the posterior distribution $q_{\vect{\phi}}$ to enhance the stability of the learning process. 
Specifically, we impose an upper bound on the KL divergence between $q_{\vect{\phi}}$ before and after the update. 
Denoting the parameter of the posterior distribution before the update as $\vect{\phi}_{\textrm{old}}$, the optimization problem in the M-step can be expressed as follows:
\begin{align}
	&\max_{q} \E_{\vect{z}\sim q} \left[  A^{\pi}(\vect{s}, \vect{a}, \vect{z})  \right] \\
	& \ \textrm{s.t.} \ \KL\left( q(\vect{z}|\vect{s}, \vect{a})||q_{\vect{\phi}_{\textrm{old}}}(\vect{z} | \vect{s}, \vect{a}) \right) \leq \epsilon_q.  \label{eq:m-step}
\end{align}
The constraint in \eqref{eq:m-step} can be viewed as a trust region constraint.
As in the optimization problem in the E-step, the solution to this problem is given in a nonparametric form as
\begin{align}
	q^*(\vect{z}|\vect{s}, \vect{a}) = \frac{1}{Z_q} q_{\vect{\phi}_{\textrm{old}}}(\vect{z}|\vect{s}, \vect{a}) \exp \left( \frac{1}{\alpha_q}  A^{\pi}(\vect{s}, \vect{a}, \vect{z}) \right),
	\label{eq:optimal_q}
\end{align}
where $Z_q$ is the partition function.
In the M-step, the parameterized posterior distribution $q_{\vect{\phi}}$ is trained to approximate the optimal posterior distribution $q^*(\vect{z}|\vect{s}, \vect{a})$ as defined in \eqref{eq:optimal_q}. 
Although the solution in \eqref{eq:optimal_q} is independent of the distributions of $\vect{s}$ and $\vect{a}$, we train our model with respect to $d^\beta(\vect{s})$ and $\pi(\vect{a}|\vect{s})$. 
This choice is motivated by our objective to obtain a solution for the problem described in \eqref{eq:problem}.

Given a dataset $\mathcal{D}$ sampled from $p^\beta(\vect{s}, \vect{a}) = d^{\beta}(\vect{s}) \beta(\vect{a}|\vect{s})$ and $\pi_{\vect{\theta}}$, 
we update $q_{\vect{\phi}}(\vect{z}|\vect{s}, \vect{a})$ to approximate the density induced by the optimal posterior distribution.
Assuming $p(\vect{z}|\vect{s})=p(\vect{z})$, we approximate $\pi(\vect{a}|\vect{s})$ as
\begin{align}
	d^{\beta}(\vect{s})\pi(\vect{a}|\vect{s})  = d^{\beta}(\vect{s})\int \pi_{\vect{\theta}}(\vect{a}|\vect{s}, \vect{z})p(\vect{z})d\vect{z}.
\end{align}
Denoting $p^{\beta, \pi}(\vect{s}, \vect{a})=d^{\beta}(\vect{s})\pi(\vect{a}|\vect{s})$, we minimize the following KL divergence to update the parameterized posterior distribution $q_{\vect{\phi}}(\vect{z}|\vect{s}, \vect{a})$:
\begin{align}
	\E_{(\vect{s}, \vect{a}) \sim p^{\beta, \pi}} \left[\KL\left( q^*(\vect{z}|\vect{s}, \vect{a}) || q_{\vect{\phi}}(\vect{z}|\vect{s}, \vect{a}) \right) \right]
\end{align}
The minimizer of the above KL divergence is given by the weighted log-likelihood: 
\begin{align}
	\mathcal{L}_{\textrm{posterior}}(\vect{\phi})= \E_{(\vect{s}, \vect{a}) \sim p^{\beta, \pi}} \E_{\vect{z}\sim q_{\vect{\phi}_{\textrm{old}}}}\left[ W_q \log q_{\vect{\phi}}(\vect{z}|\vect{s}, \vect{a}) \right],
	\label{eq:posterior_weighted_ml}
\end{align}
where the weight $W_q(\vect{s}, \vect{a}, \vect{z})$ is given by
\begin{align}
	W_q(\vect{s}, \vect{a}, \vect{z}) = \frac{1}{Z_q} \exp \left(\frac{1}{\alpha_q} A^\pi(\vect{s}, \vect{a}, \vect{z}) \right).
\end{align}
While the form of the solution in \eqref{eq:posterior_weighted_ml} is analogous to that in \eqref{eq:awr_latent}, 
the temperature parameters $\alpha_\pi$ and $\alpha_q$ are different because they are separately derived from constraints in \eqref{eq:KL_RL_prob_latent} and \eqref{eq:m-step}.
Thus,  $\alpha_\pi$ and $\alpha_q$ should be tuned separately in practice.

In addition, to assure the consistency between the trained posterior distribution and policy, we also train the posterior distribution so as to approximate the density induced by the optimal policy in \eqref{eq:E-step-nonpara}.
For this purpose, we introduce a likelihood $p_{\vect{\psi}}(\vect{s}, \vect{a}|\vect{z})$ parameterized with a vector $\vect{\psi}$.

Given a posterior distribution $q(\vect{z}|\vect{s}, \vect{a})$, the density $ p^*(\vect{s}, \vect{a})$ induced by the optimal policy in \eqref{eq:E-step-nonpara} is given by
\begin{align}
	& p^*(\vect{s}, \vect{a})  = d^{\beta}(\vect{s}) \pi^*(\vect{a}|\vect{s}) \\
	& = \int d^{\beta}(\vect{s}) )p(\vect{z} |\vect{s}) \pi^*(\vect{a}|\vect{s},\vect{z}) d\vect{z} \\
	& = \int d^{\beta}(\vect{s}) p(\vect{z} |\vect{s}) \beta(\vect{a}|\vect{s}, \vect{z})  \frac{1}{Z_\pi}  \exp \left( \frac{1}{\alpha_\pi}  A^{\pi}(\vect{s}, \vect{a}, \vect{z}) \right) d\vect{z} \\
	& = \frac{1}{Z_\pi} \int d^{\beta}(\vect{s})  \beta(\vect{a}|\vect{s}) q(\vect{z} |\vect{s}, \vect{a})  \exp \left( \frac{1}{\alpha_\pi}  A^{\pi}(\vect{s}, \vect{a}, \vect{z}) \right) d\vect{z}
	\label{eq:optimal_p}
\end{align}
The result in \eqref{eq:optimal_p} indicates that the  density $ p^*(\vect{s}, \vect{a})$ can be approximated by 1)~sampling the latent variable $\vect{z}$ with $q$, 2)~set the importance weight with $\exp \left( \frac{1}{\alpha_\pi}  A^{\pi}(\vect{s}, \vect{a}, \vect{z}) \right)$, and 3) marginalize over the latent variable $\vect{z}$.


%
To train the parameterized posterior distribution and likelihood, we use the variational lower bound used in variational autoencoder (VAE)~\cite{Kingma14}, which is given by
\begin{align}
	\log p(\vect{s}_i, \vect{a}_i)  
	\geq & \ell_{\vect{\phi}, \vect{\psi}}(\vect{s}_i, \vect{a}_i) \\
	= & -\KL \left( q_{\vect{\phi}}(\vect{z}|\vect{s}_i, \vect{a}_i)||p(\vect{z}) \right) \nonumber \\
	& + \E_{\vect{z}\sim q_{\vect{\phi}}} \left[ \log p_{\vect{\psi}}(\vect{s}_i, \vect{a}_i|\vect{z}) \right].
	\label{eq:vae}
\end{align}

Combining \eqref{eq:optimal_p} and \eqref{eq:vae}, we maximize the following objective:
\begin{align}
	& \E_{(\vect{s}, \vect{a})\sim p^*}\left[ \log p(\vect{s}, \vect{a}) \right]\nonumber\\
	& = \E_{(\vect{s}, \vect{a})\sim d^\beta, \beta} \E_{\vect{z}\sim q} \left[   W_\pi(\vect{s}, \vect{a}, \vect{z}) \log p(\vect{s}, \vect{a}) \right] \\
	& \geq \E_{(\vect{s}, \vect{a})\sim d^\beta, \beta} \E_{\vect{z}\sim q} \left[ W_\pi(\vect{s}, \vect{a}, \vect{z}) \ell_{\vect{\phi}, \vect{\psi}}(\vect{s}, \vect{a})  \right] \\
	& \approx \frac{1}{N\cdot N_z} \sum^N_{i=1} \sum^{N_z}_{j=1} W_\pi(\vect{s}_i, \vect{a}_i, \vect{z}_j) \ell_{\vect{\phi}, \vect{\psi}}(\vect{s}_i, \vect{a}_i) \\
	& \coloneqq \mathcal{L}_{\textrm{w-vae}}(\vect{\phi}, \vect{\psi})
	\label{eq:weighted_vae}
\end{align}
where $W_\pi(\vect{s}, \vect{a}, \vect{z})$ is given as in \eqref{eq:weight_pi}.

Based on the result in \eqref{eq:posterior_weighted_ml} and \eqref{eq:weighted_vae}, we maximize the following objective function in the M-step:
\begin{align}
	\mathcal{L}_{\textrm{M-step}}(\vect{\phi}, \vect{\psi})= \mathcal{L}_{\textrm{posterior}}(\vect{\phi}) +  \mathcal{L}_{\textrm{w-vae}}(\vect{\phi}, \vect{\psi}).
	\label{eq:m-step-update}
\end{align}

\subsection{Learning the Latent Skill Space via Mutual Information Maximization}
\label{sec:infomax}
Herein, we discuss how to maximize the mutual information term in \eqref{eq:problem}.
For this purpose, we utilize the variational lower bound of the mutual information. 
We consider mutual information $I(\vect{z}; \vect{s}, \vect{a})$ between the latent variable $\vect{z}$ and a state-action pair $(\vect{s}, \vect{a})$ as
\begin{align}
	I(\vect{z}; \vect{s}, \vect{a}) = \iiint p(\vect{s}, \vect{a}, \vect{z}) \log \frac{p(\vect{s}, \vect{a},\vect{z})}{p(\vect{s}, \vect{a})p(\vect{z})} d\vect{z}d\vect{a}d\vect{s}.
\end{align}
The variational lower bound of $I(\vect{z}; \vect{s}, \vect{a})$ is then given as follows~\cite{Barber03,Osa22}:
\begin{align}
	I(\vect{z}; \vect{s}, \vect{a}) \geq  \E_{(\vect{s}, \vect{a}, \vect{z}) \sim p} \left[ \log  q(\vect{z} | \vect{s}, \vect{a}) \right] +  H(\vect{z}),
	\label{eq:info_bound}
\end{align}
where $H(\vect{z})$ is the entropy of the latent variable $\vect{z}$.
As $H(\vect{z})$  is independent of the policy, encoder, and decoder parameters, 
only the first term of the right-hand side in \eqref{eq:info_bound} must be maximized.

As we maximize the lower bound with respect to $p^\pi(\vect{s}, \vect{a}, \vect{z})= d^\beta(\vect{s})p(\vect{z}|\vect{s})\pi(\vect{a}|\vect{s}, \vect{z})$, 
we maximize the following term:
\begin{align}
	& \mathcal{L}_{\textrm{info}}(\vect{\theta}, \vect{\phi}) = \E_{(\vect{s}, \vect{a}, \vect{z}) \sim p^\pi} \left[ \log  q_{\vect{\phi}}(\vect{z} | \vect{s}, \vect{a}) \right] \nonumber \\
	& = \int d^{\beta}(\vect{s}) p(\vect{z}|\vect{s})\pi_{\vect{\theta}}(\vect{a}|\vect{s}, \vect{z}) \log  q(\vect{z} | \vect{s}, \vect{a}) d\vect{z}d\vect{a}d\vect{s} \label{eq:inf_loss}
\end{align}

As shown by previous studies~\cite{Kumar20a,Osa22}, maximizing the  variational lower bound of mutual information $I(\vect{z};\vect{s}, \vect{a})$ encourages the latent-conditioned policy to generate different actions for different values of the latent variable.
Consequently, the diversity of actions is enhanced by maximizing this term.
Additionally, when maximizing $\E_{\tilde{\vect{a}}\sim \pi_{\vect{\theta}}}\left[ \log q_{\vect{\phi}}(\vect{z} | \vect{s}, \tilde{\vect{a}}) \right]$, the encoder $q_{\vect{\phi}}(\vect{z} | \vect{s}, \vect{a})$ recovers the value of the latent variable from the state and action generated by the latent-conditioned policy $\pi_{\vect{\theta}}(\vect{a}|\vect{s}, \vect{z})$.
Thus, the encoder is updated to learn latent representations that are consistent with the behavior of the latent-conditioned policy.

Combining the objectives for the M-step and mutual information maximization, we jointly update the policy, posterior distribution, and likelihood by maximizing the following objective function with the regularization based on mutual information maximization:
\begin{align}
	 \mathcal{L}_{\textrm{M-info}}(\vect{\phi}, \vect{\psi}, \vect{\theta}) = \mathcal{L}_{\textrm{M-step}}(\vect{\phi}, \vect{\psi}) + \lambda \mathcal{L}_{\textrm{info}}(\vect{\theta}, \vect{\phi}) ,
	\label{eq:vae_info}
\end{align}
where $\lambda$ is a constant that balances the two terms.

\subsection{Practical Algorithm}

\begin{algorithm}[tb]
	\caption{Learning Diverse Behaviors in Offline RL (DiveOff)}
	\label{alg:diveoff}
	\begin{algorithmic}
		\STATE {\bfseries Input:} dataset $\mathcal{D}=\{(\vect{s}_i, \vect{a}_i, r_i, \vect{s}_i')  \}^N_{i=1}$
		\STATE Initialize the policy $\pi_{\vect{\theta}}$, critic $Q_{\vect{w}_j}$ for $j=1, 2$, posterior $q_{\vect{\phi}}(\vect{z}|\vect{s}, \vect{a})$, and likelihood $p_{\vect{\psi}}(\vect{s}, \vect{a}|\vect{z})$
		\STATE Train the posterior $q_{\vect{\phi}}(\vect{z}|\vect{s}, \vect{a})$, and likelihood $p_{\vect{\psi}}(\vect{s}, \vect{a}|\vect{z})$ with $\mathcal{D}$ using the standard VAE loss
		\FOR{$t=1$ {\bfseries to} $T$}
		\STATE Sample a minibatch $\{(\vect{s}_i, \vect{a}_i, \vect{s}_i', r_i )\}^N_{i=1}$ from $\mathcal{D}$
		\STATE Sample the latent variable $\vect{z}_i \sim q_{\vect{\phi}}(\vect{z}|\vect{s}_i, \vect{a}_i)$
		\STATE Compute the target value:
		\STATE \ \ \ $y_i = r + \gamma \min_{j=1,2} Q_{\vect{w}_j}(\vect{s}_i, \vect{a}', \vect{z}_i ) $,  
		\STATE \ \ \ where $\vect{a}' \sim \pi_{\vect{\theta}}(\vect{a}'|\vect{s}'_i, \vect{z}_i)$
		\STATE Update the critic:
		\STATE \ \ \ $N^{-1} \sum^N_{i=1} \left( y_i - Q_{\vect{w}_j}(\vect{s}_i, \vect{a}_i, \vect{z}) \right)^2$ for $j=1,2$
		\IF{$t \ \textrm{mod} \ d_{\textrm{interval}} = 0$}
		\STATE Update the policy by maximizing $\mathcal{L}_{\textrm{E-step}}(\vect{\theta})$ in \eqref{eq:awr_latent}
		\STATE Update the policy, posterior, and likelihood by maximizing $\mathcal{L}_{\textrm{M-info}}(\vect{\phi}, \vect{\psi}, \vect{\theta})$ in \eqref{eq:vae_info}
		\ENDIF
		\ENDFOR
	\end{algorithmic}
\end{algorithm}

The proposed algorithm, referred to as Learning Diverse Behaviors in Offline RL~(DiveOff), is summarized in Algorithm~\ref{alg:diveoff}.
We initialize the posterior $q_{\vect{\phi}}$ and likelihood $p_{\vect{\psi}}$ using the standard VAE objective function.
We then iterate the E-step and M-step described in the previous section to update the policy and posterior distribution.
Whereas Algorithm~\ref{alg:diveoff} is described assuming that the critic is trained with double-clipped Q-learning~\cite{Fujimoto18}, 
alternative methods can be employed, such as expectile regression as in IQL.

\section{Experiments}
\label{sec:exp}
To analyze the performance of the proposed method, we initially assess it on a two-dimensional toy task. 
Following this, we conduct evaluations on locomotion tasks in Mujoco, utilizing our original datasets that encompass a wide range of diverse behaviors.

\subsection{Evaluation on Toy Task}
\begin{figure}[tb]
	\vspace{-0.3cm}
	\centering
	\subfigure[Task setting and data samples.]{\includegraphics[width=0.32\columnwidth]{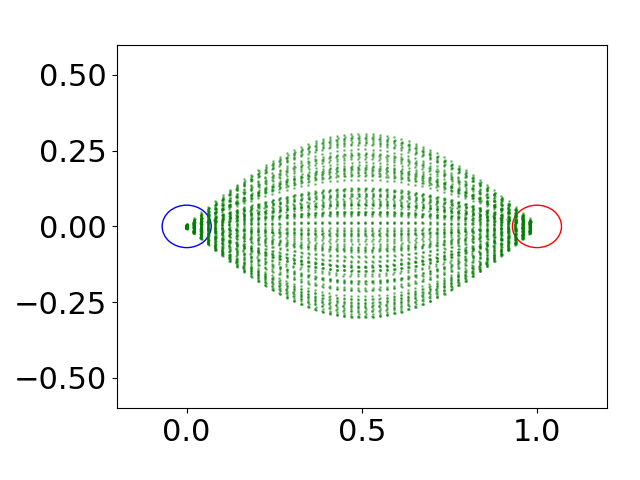}}
	\hfil
	\subfigure[Trajectories generated by the proposed method.]{\includegraphics[width=0.32\columnwidth]{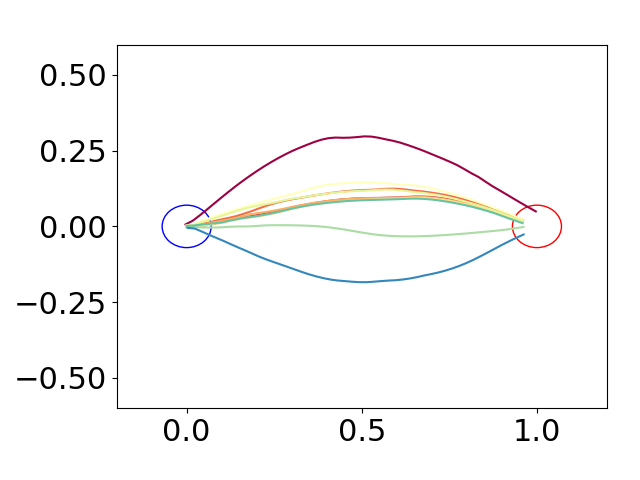}}
	\hfil
	\subfigure[The latent variable learned with the proposed method.]{\includegraphics[width=0.32\columnwidth]{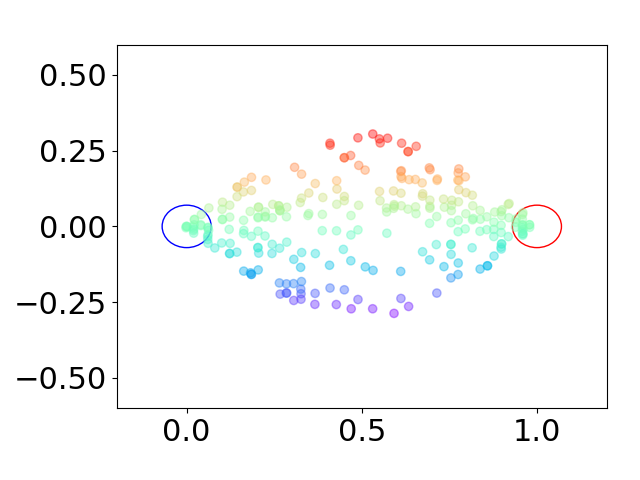}}
	\vspace{-0.3cm}
	\caption{Path planning task on two-dimensional space. }
	\label{fig:toy-task}
	\vspace{-0.3cm}
\end{figure}

To visualize the behavior of the proposed method in a simple task, we evaluated the performance of our method in a path planning problem on two dimensional space.
In this toy task, the agent is represented as a point mass, the state is the position of the point mass in two-dimensional space, and the action is the small displacement of the point mass. 
The blue circle denotes the starting position, and the agent receives the reward if it reaches the red circle.
Figure~\ref{fig:toy-task}(a) shows the data samples in the dataset. Figure~\ref{fig:toy-task}(b) shows the multiple solutions learned by the proposed method.
The proposed method may not necessarily cover all solutions, but it can be confirmed that multiple solutions have been discovered by the proposed approach.

Figure~\ref{fig:toy-task}(c) shows the latent variable learned by the proposed method, which learns the latent variable consistent with trajectories or type of solutions. 
In (c), the color indicates the mean of the latent variable, $\vect{\mu}$, where the posterior distribution $q(\vect{z}|\vect{s}, \vect{a})$ is given by Gaussian $\mathcal{N}(\vect{\mu}, \vect{\sigma})$. 
The colors are separately assigned along the vertical axis, because the vertical positions characterize the behavior of the latent-conditioned policy. 
This result indicates that the proposed method learns the latent representations that are consistent with the types of solutions.

\subsection{Performance on D4RL Tasks with Diverse Behaviors}

\subsubsection{Datasets with Diverse Behaviors}
We constructed datasets that contain diverse locomotion behaviors based on the D4RL framework.
To emphasize the existence of diverse solutions, we modified the velocity term of the reward function from that in the original tasks in the D4RL framework,
as in \cite{Kumar20}: 
\begin{align}
	r_{\textrm{vel}} = \min\big(  (x_{t} - x_{t-1}) / \Delta t, v_{\textrm{max}} \big),
\end{align}
where $x_t$ represents the horizontal position of the agent at time $t$, and $v_{\textrm{max}}$ is a constant term that defines the upper bound of $r_{\textrm{vel}}$.
We refer to tasks with the above reward term as xxVel, e.g., Walker2dVel. 
To construct datasets with diverse locomotion behaviors, a latent-conditioned policy was trained with the method presented in~\cite{Osa22}, and we collected samples by executing the trained policy with random values of the latent variable.
We repeated this process five times with different random seeds, and aggregated the collected samples into a dataset. 
Following the convention of the D4RL framework, we prepared four different types of datasets based on the policy level used to collect samples: ``expert,'' ``medium-expert,'' ``medium,'' and ``medium-replay.''
We describe the tasks in detail in Appendix~\ref{app:dataset}.
Codes and videos are available in \url{https://takaosa.github.io/project_diveoff.html}.

\begin{table*}[tb]
	\vskip -0.15in
		\caption{ Normalized scores of methods that train latent-conditioned policies.
			WK = Walker2dVel, HP = HopperVel, HC = HalfcheetahVel, AN=AntVel. Best results are denoted in bold. 
			}
		\label{tbl:diverse_results_score}
		\vskip -0.15in
		\begin{center}
			\begin{small}
				\begin{sc}
					\begin{tabular}{clr|rrrrrrr}
						\toprule
						& & TD3+BC & CLUE &\parbox[c]{1.3cm}{\centering AWAC-L + VAE } &  \parbox[c]{1.3cm}{\centering IQL-L + VAE } & \parbox[c]{1.3cm}{\centering AWAC-L + VAE w/ LSTM} & \parbox[c]{1.3cm}{\centering AWAC-L + VAE + DIAYN} & \parbox[c]{1.3cm}{\centering  DiveOff w/  VAE (ours) }  & \parbox[c]{1.3cm}{\centering DiveOff (ours) }   \\
						\midrule
						\multirow{4}{*}{\rotatebox[origin=c]{90}{\parbox[c]{1.5cm}{\centering  div.-exp.}}} 
						& WK & 75.5$\pm$31.9 & 59.1$\pm$6.7 &79.9$\pm$35.1 &  43.5$\pm$38.7 & 82.2$\pm$26.7 & 86.7$\pm$15.5 & 77.2 32.7 & \textbf{93.3$\pm$9.9}\\
						& HP & 94.6$\pm$11.5 & 39.5$\pm$17.8 &\textbf{98.6$\pm$3.0} & 4.9$\pm$3.8 & 94.0$\pm$3.8 & \textbf{98.0$\pm$3.3} & \textbf{99.7$\pm$0.6} & 81.2 33.8\\
						& HC & 97.6$\pm$0.2 & 95.8$\pm$0.8 &\textbf{96.5$\pm$0.4} & 94.0$\pm$5.3 & \textbf{96.6$\pm$0.2} & 96.3$\pm$0.5 & \textbf{97.0$\pm$0.2} & \textbf{96.8$\pm$0.3}\\
						& AN & 98.5$\pm$2.8 & 78.2$\pm$7.8 &94.5$\pm$1.1 & \textbf{97.2$\pm$3.4} & \textbf{95.4$\pm$0.4} &  86.8$\pm$5.4 & \textbf{95.8$\pm$0.8} & \textbf{95.4$\pm$1.2} \\
						\midrule
						\multirow{4}{*}{\rotatebox[origin=c]{90}{\parbox[c]{1.5cm}{\centering  div.-exp.-med.}}} 
						& WK & 96.5$\pm$7.0 & 54.7$\pm$15.3 &93.7$\pm$9.1 & 42.0$\pm$27.4 & 62.9$\pm$6.9 & \textbf{97.1$\pm$2.2} & 95.8$\pm$5.9 & 80.5 17.7 \\
						& HP & 92.1$\pm$14.0 & 74.8$\pm$22.8 &94.8$\pm$7.2 & 89.3$\pm$23.4 & 79.7$\pm$24.3 & 78.7$\pm$36.5 & 94.8$\pm$9.2 & \textbf{98.4$\pm$4.7} \\
						& HC & 97.1$\pm$4.2 & \textbf{96.2$\pm$0.4} &\textbf{96.0$\pm$0.5} &  91.8$\pm$14.2 & \textbf{96.9$\pm$0.2} & \textbf{96.2$\pm$0.7} & \textbf{96.3$\pm$0.2} &  \textbf{96.1$\pm$0.6}\\
						& AN & 96.2$\pm$3.8 & 79.2$\pm$10.6 &84.8$\pm$6.2 & \textbf{95.3$\pm$3.0} & 87.4$\pm$3.8 & 82.1$\pm$6.3 & 89.8$\pm$0.6 & 88.5$\pm$1.9 \\
						\midrule
						\multirow{4}{*}{\rotatebox[origin=c]{90}{\parbox[c]{1.5cm}{\centering  div.-med.}}} 
						& WK & 77.0$\pm$13.7 & 47.4$\pm$14.2 &\textbf{84.6$\pm$4.9} & 74.3$\pm$23.2 & 57.3$\pm$15.9 & 74.8$\pm$11.1 & 79.1$\pm$19.3 &  67.9$\pm$17.4\\
						& HP & 88.1$\pm$9.0 & 65.0$\pm$7.2 &93.1$\pm$5.3 & 87.1$\pm$9.1 & 82.2$\pm$15.6 & 92.4$\pm$6.3 & \textbf{94.5$\pm$3.4} & 83.1$\pm$18.0 \\
						& HC & 92.3$\pm$2.3 & 92.3$\pm$3.0 &\textbf{93.4$\pm$0.3} & \textbf{93.5$\pm$4.0} & \textbf{93.0$\pm$0.4} & \textbf{93.3$\pm$0.6} & 93.3$\pm$0.4 &  \textbf{93.2$\pm$0.5}\\
						& AN & 64.3$\pm$64.1 & 66.3$\pm$5.4 &80.5$\pm$2.4 & \textbf{92.7$\pm$0.9} & 76.5$\pm$5.1 & 75.3$\pm$5.7 &  81.8$\pm$2.6 & 81.1$\pm$2.3 \\
						\midrule
						\multirow{4}{*}{\rotatebox[origin=c]{90}{\parbox[c]{1.5cm}{\centering  div.-med.-rep.}}} 
						& WK & 80.6$\pm$28.4 & 17.5$\pm$13.0 &26.8$\pm$9.3 & \textbf{88.3$\pm$9.7} & 43.5$\pm$26.8 & 51.4$\pm$5.9 & 27.5$\pm$29.0 &  39.3$\pm$11.6\\
						& HP & 67.9$\pm$23.1 & 93.3$\pm$7.4 &96.6$\pm$5.8 & \textbf{101.0$\pm$0.1} & 6.6$\pm$3.9 & 58.9$\pm$42.4 & 99.1$\pm$3.9 &  \textbf{101.0$\pm$0.1}\\
						& HC & 93.2$\pm$0.6 & 89.9$\pm$2.5 &\textbf{91.1$\pm$3.0} &   \textbf{92.7$\pm$0.7} & \textbf{92.2$\pm$0.7} & 89.3$\pm$3.8 & 80.4$\pm$25.0 &  89.8$\pm$6.4\\
						& AN & 91.9$\pm$3.2 & 18.5$\pm$3.8 &34.6$\pm$2.3 & \textbf{94.6$\pm$0.8} & 35.5$\pm$1.6 & 34.7$\pm$3.0 & 33.9$\pm$2.3 &  32.9$\pm$2.9\\
						\bottomrule
					\end{tabular}
				\end{sc}
			\end{small}
		\end{center}
		\vskip -0.1in
\end{table*}
\begin{table*}[tb]
		\caption{ Diversity scores of methods that train latent-conditioned policies.
			WK = Walker2dVel, HP = HopperVel, HC = HalfcheetahVel, AN=AntVel. Best results are denoted in bold. The results for cases where the D4RL scores fall below 20.0 compared to the best score are presented in gray text. The results for cases where the D4RL scores are the best among the compared methods are presented with underlines.
			}
		\label{tbl:diverse_results_diversity}
		\vskip -0.15in
		\begin{center}
			\begin{small}
				\begin{sc}
					\begin{tabular}{clrrrrrrr}
						\toprule
						& & CLUE & \parbox[c]{1.3cm}{\centering AWAC-L + VAE } &  \parbox[c]{1.3cm}{\centering IQL-L + VAE } & \parbox[c]{1.3cm}{\centering AWAC-L + VAE w/ LSTM} & \parbox[c]{1.3cm}{\centering AWAC-L + VAE + DIAYN} & \parbox[c]{1.3cm}{\centering  DiveOff + VAE (ours) }  & \parbox[c]{1.3cm}{\centering DiveOff (ours) }   \\
						\midrule
						\multirow{4}{*}{\rotatebox[origin=c]{90}{\parbox[c]{1.5cm}{\centering  div.-exp.}}} 
						& WK & \gray{0.95$\pm$0.10} &  0.13$\pm$0.12 & \gray{0.20$\pm$0.40} & 0.35$\pm$0.36 & \textbf{0.48$\pm$0.33} & 0.18$\pm$0.34 & \underline{0.22$\pm$0.23}\\
						& HP & \gray{0.99$\pm$0.01} &  0.65$\pm$0.33 &  \gray{0.06$\pm$0.08} & 0.55$\pm$0.28 & 0.54$\pm$0.33 & 0.28$\pm$0.32 & \textbf{0.88$\pm$0.11}\\
						& HC & \textbf{0.99$\pm$0.01} &  \textbf{0.96$\pm$0.06} &  0.20$\pm$0.15 & 0.62$\pm$0.38 & 0.90$\pm$0.16 & 0.85$\pm$0.11 & \underline{0.77$\pm$0.24}\\
						& AN & \textbf{0.71$\pm$0.32}&  \textbf{0.20$\pm$0.34} &  0.02$\pm$0.02 & 0.19$\pm$0.32 & 0.03$\pm$0.03 & 0.06$\pm$0.09 & \underline{0.09$\pm$0.11} \\
						\midrule
						\multirow{4}{*}{\rotatebox[origin=c]{90}{\parbox[c]{1.5cm}{\centering  div.-exp.-med.}}} 
						& WK & \gray{0.98$\pm$0.02} &  0.38$\pm$0.39 &  \gray{0.432$\pm$0.38} & \gray{0.81$\pm$0.23} & 0.21$\pm$0.22 & 0.29$\pm$0.26 & \textbf{0.75$\pm$0.20} \\
						& HP & \gray{0.99$\pm$0.01} &  0.05$\pm$0.07 &  0.36$\pm$0.39 & 0.23$\pm$0.22 & \gray{0.21$\pm$0.40} & 0.31$\pm$0.36&  \underline{\textbf{0.88$\pm$00.19}}\\
						& HC & \textbf{0.99$\pm$0.01} &  0.72$\pm$0.20 & 0.40$\pm$0.48 & \underline{0.81$\pm$0.22} & 0.62$\pm$0.26 & 0.65$\pm$0.30 & \underline{\textbf{0.96$\pm$0.04}} \\
						& AN & 0.42$\pm$0.25 &  0.38$\pm$0.27 &  \underline{0.04$\pm$0.05} & 0.18$\pm$0.28 & 0.407$\pm$0.22 & 0.38$\pm$0.29 & \textbf{0.61$\pm$0.27} \\
						\midrule
						\multirow{4}{*}{\rotatebox[origin=c]{90}{\parbox[c]{1.5cm}{\centering  div.-med.}}} 
						& WK & \gray{0.99$\pm$0.01} &  0.25$\pm$0.25 &  \textbf{0.43$\pm$0.38} & \gray{0.44$\pm$0.29} & 0.31$\pm$0.25 & 0.34$\pm$0.37 & \textbf{0.36$\pm$0.33} \\
						& HP & \gray{0.99$\pm$0.01} &  0.46$\pm$0.28 & \textbf{0.79$\pm$0.25} & 0.60$\pm$0.34 & 0.61$\pm$0.36 & 0.45$\pm$0.33 & \textbf{0.79$\pm$0.34} \\
						& HC & \textbf{0.99$\pm$0.01} &  0.52$\pm$0.25 &  \underline{0.05$\pm$0.04} & 0.73$\pm$0.25 & 0.57$\pm$0.32 & 0.75$\pm$0.22 & \underline{0.81$\pm$0.13} \\
						& AN & \gray{0.84$\pm$0.17} &  0.56$\pm$0.35 & \underline{0.02$\pm$0.04} & \gray{0.15$\pm$0.23} & 0.56$\pm$0.31 & 0.38$\pm$0.29 & \textbf{0.65$\pm$0.26} \\
						\midrule
						\multirow{4}{*}{\rotatebox[origin=c]{90}{\parbox[c]{1.5cm}{\centering  div.-med.-rep.}}} 
						& WK & 0.81$\pm$0.33 &  \gray{0.00$\pm$0.00} &  \underline{\textbf{0.36$\pm$0.40}} & \gray{0.39$\pm$0.48} & \gray{0.32$\pm$0.39} & \gray{0.01$\pm$0.02} & \gray{0.01$\pm$0.01} \\
						& HP & \textbf{0.99$\pm$0.01} &  0.00$\pm$0.02 & \underline{0.00$\pm$0.00} & \gray{0.46$\pm$0.39} & \gray{0.78$\pm$0.27} & 0.00$\pm$0.00 &\underline{0.02$\pm$0.03} \\
						& HC & \textbf{0.98$\pm$0.02} &  0.01$\pm$0.03 &  \underline{0.14$\pm$0.16} & 0.04$\pm$0.08 & 0.01$\pm$0.02 & 0.00$\pm$0.00 & 0.00$\pm$0.00 \\
						& AN & 0.33$\pm$0.41 &  \gray{0.24$\pm$0.21} &  \underline{0.0$\pm$0.0} & \gray{0.43$\pm$0.47} &  \gray{0.40$\pm$0.49} &  \gray{0.42$\pm$0.48} & \gray{0.06$\pm$0.12} \\
						\bottomrule
					\end{tabular}
				\end{sc}
			\end{small}
		\end{center}
		\vskip -0.15in
\end{table*}

\subsubsection{Comparison with Baseline Methods}
As a baseline method, we evaluated CLUE, which was proposed by \citet{Liu23}.
To investigate the effect of the components of the proposed algorithm, we devised several methods for comparison ourselves and compared them with the proposed method.
The proposed policy update rule presented in \eqref{eq:weight_pi} can be combined with other methodologies for learning the latent representations.
We refer to baseline methods based on  \eqref{eq:weight_pi} as AWAC-L.
As a baseline method for learning the latent representations, we evaluated methods based on VAE.
In a baseline method using VAE, VAE was trained with a given dataset $\mathcal{D}$.
The latent variable estimated by VAE is augmented with the state as $\tilde{\vect{s}} = [\vect{s}, \vect{z}]$, and the policy that takes in $\tilde{\vect{s}}$ was trained with AWAC-L.
We refer to this baseline method as AWAC-L+VAE.
We also evaluated a variant of AWAC-L+VAE where we trained the critic with the expectile regression as in IQL~\cite{Kostrikov22}.
We refer to this variant of AWAC-L+VAE as IQL-L+VAE.
In our implementation of baseline methods except IQL-L, we used the double-clipped Q-learning as in TD3+BC~\cite{Fujimoto21}.
In our implementation of AWAC-L+VAE and IQL-L+VAE, VAE was implemented with two hidden fully-connected layers.

As a variant of the AWAC-L+VAE, we also evaluated VAE with LSTM layers, which can consider temporal sequences for learning the latent representations.
We refer to this variant of AWAC-L+VAE as AWAC-L+VAE with LSTM.
In OPAL proposed by \citet{Ajay21}, the latent variables are learned with a recurrent neural network using a VAE-like loss function, and the latent-conditioned policy learned. 
In a sense that the latent variable encodes temporal information, AWAC-L+VAE with LSTM is similar to OPAL.
The state normalization was employed in all the methods, including the proposed methods.

In the context of online RL, maximizing the mutual information between the state and latent variables is a prevalent approach to obtain diverse behaviors.
As a baseline, we evaluated a variant of AWAC that maximizes the mutual information term in addition to the task reward as in \cite{Kumar20a,Sharma20}.
When introducing a model that approximates the posterior distribution $q(\vect{z}|\vect{s})$, we can consider the auxiliary reward term given by
\begin{align}
	r_{\textrm{info}} = \log q_{\vect{\psi}}(\vect{z}|\vect{s}),
	\label{eq:diayn_reward}
\end{align}
where $q_{\vect{\psi}}(\vect{z}|\vect{s})$is the approximated posterior distribution. 
The auxiliary reward term in \eqref{eq:diayn_reward} was used in DIAYN~\cite{Eysenbach19}, and it was later used by~\citet{Kumar20a,Sharma20}.
We refer to the variant of AWAC-L+VAE that maximizes the reward term in \eqref{eq:diayn_reward} as AWAC-L+VAE+DIAYN.

To assess the impact of the components in our proposed method, we conducted an evaluation on a variant of DiveOff. 
In this variant, the posterior distribution was trained as part of a VAE, and the objective function in \eqref{eq:m-step-update} was replaced with the standard VAE loss. 
We denote this variant as DiveOff+VAE. In DiveOff+VAE, the posterior distribution and the policy are alternately updated.
SORL proposed by \citet{Mao24} iterates the clustering and policy improvement, without advantage-based weights in the clustering.
In this sense, DiveOff+VAE can be interpreted as a variant of SORL.
To quantify the diversity of the learned solutions, we employed the diversity metric used by \citet{Parker-Holder20}.
We provide details on how to compute the diversity score in Appendix~\ref{app:diversity_score}.
In the following, we report the mean and standard deviation over the 10 test episodes and five seeds unless otherwise stated.
The value of the latent variable was sampled from the uniform distribution $U(-1, 1)$ at the beginning of the test episodes.
The latent variable was continuous and two-dimensional in this experiment.

The normalized scores attained by the proposed method's variants are listed Table~\ref{tbl:diverse_results_score}.
Our algorithm consistently achieved high d4rl scores, demonstrating the stability of the learning process. 
A comparison with baseline methods that learn a single solution is provided in Appendix~\ref{app:comp_single}.
The results of a variant of DiveOff based on the expectile regression as in IQL is also provided in Appendix~\ref{app:diveoff_iql}.

As outlined in Table~\ref{tbl:diverse_results_diversity}, our algorithm demonstrated a high diversity score. 
Notably, our method stands out as the sole approach capable of achieving both a high diversity score and a high d4rl score across tasks. 
Furthermore, the contrast between DiveOff+VAE and DiveOff underscores the efficacy of our proposed methodology in learning latent skill representations, thereby enhancing the diversity of the acquired policy.

While it was demonstrated that diverse behaviors can be learned with CLUE, the performance on the training tasks were not stable, compared with DiveOff.
It is worth noting that CLUE was originally developed to leverage expert demonstrations in offline RL, and it is not specifically designed to discover multiple solutions from a single task in offline RL. 
When CLUE is applied to unsupervised RL, \citet{Liu23} proposed to use the discrete latent representation, learned by a clustering algorithm such as K-means. 
However, unlike our method, important information such as the Q-values is not incorporated with such latent representations. Therefore, it is natural that DiveOff outperforms CLUE in our problem setting.

The comparison between AWAC-L+VAE+DIAYN and DiveOff with VAE in Tables 1 and 2 indicates the effect of strategies for maximizing mutual information. 
DIAYN is a prevalent method to encourage the diversity of behaviors in online RL, maximizing an auxiliary reward given by the lower bound of the mutual information between the state and latent variables. The results indicate that our strategy proposed in 
Section~\ref{sec:infomax} is more suitable for our framework than DIAYN.

The comparison between AWAC-L+VAE and AWAC-L+VAE with LSTM shows that the use of LSTM in the posterior distribution does not improve performance in our problem setting. 
While AWAC-L+VAE with LSTM demonstrated reasonable performance with the "diverse-expert" datasets, the performance with "diverse-medium" and "diverse-medium-replay" datasets was relatively worse. 
The ``diverse-medium'' and ``diverse-medium-replay'' datasets do not contain expert behaviors, and it is considered that the ``stitching'' capability is more crucial for learning from these datasets.

Notably, the methods evaluated in this study often failed to extract diverse behaviors from the ``diverse-medium-replay'' datasets. 
While it is believed that the ``diverse-medium-replay'' data may have the most diverse actions, the diversity of successful states is considered to be much less than that of the ``diverse-expert'' dataset. 
If the successful states are not diverse, the actions within those states are also not diverse. 
It is considered that this characteristic of the "diverse-medium-replay" datasets is attributed to the failure observed in these datasets.

\begin{figure}[tb]
	\centering
	\includegraphics[width=\columnwidth]{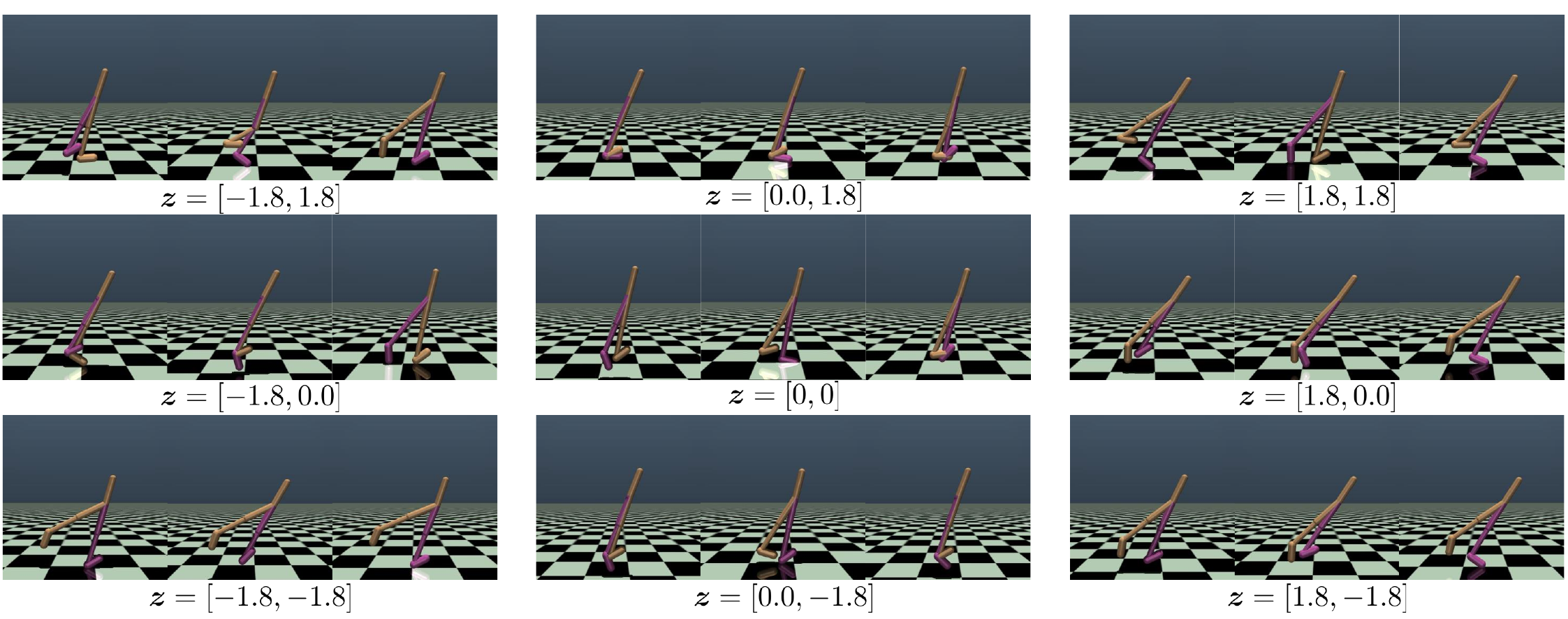}
	\vspace{-0.3cm}
	\caption{Sequential snapshots of locomotion behaviors obtained by DiveOff on the walker2dvel-diverse-expert task. }
	\label{fig:AWAC_V_W_diverse_expert_walker}
\end{figure}

As a qualitative result, Figure~\ref{fig:AWAC_V_W_diverse_expert_walker} visualizes   the locomotion behaviors obtained by DiveOff on the walker2dvel-diverse-expert tasks.
Here, the agent performed one-leg hopping when $\vect{z}=[-1.8, -1.8]$, and walked with two legs when $\vect{z}=[1.8, 1.8]$.
As the policy is conditioned on a continuous latent variable, we can continuously change the locomotion behavior by manipulating the latent variable's value. 

A limitation of this study is that the learned behavior varies with different random seeds. 
This is a common characteristic in existing deep reinforcement learning algorithms. 
However, we posit that addressing this issue could be feasible if it were possible to learn all learnable behaviors. 
Consequently, there is room for future development of algorithms with the potential to acquire a more diverse range of behaviors.

\begin{table*}[tb]
	\caption{ Results of few-shot adaptation with the hopper agent.  }
	\label{tbl:fewshot_results}
	\begin{center}
		\begin{small}
			\begin{sc}
				\begin{tabular}{cclrrrr}
					\toprule
					 & \parbox[c]{1.2cm}{\centering Train. Dataset } & \parbox[c]{1.2cm}{\centering Test Agent } &  \parbox[c]{1.3cm}{\centering AWAC-L + VAE } & \parbox[c]{1.3cm}{\centering AWAC-L + VAE + DIAYN} & \parbox[c]{1.3cm}{\centering  DiveOff w/ VAE (ours) }  & \parbox[c]{1.3cm}{\centering DiveOff (ours) }   \\
										\midrule
										 &\multirow{2}{*}{\parbox[c]{1.5cm}{\centering div.-exp.}} 
										& LowKnee & 766.4$\pm$631.4 & 817.7$\pm$623.8 & \textbf{1374.76$\pm$663.6} & 764.7$\pm$330.1\\
										& & LongHead & 1139.7$\pm$468.0 & \textbf{1440.6$\pm$563.7} & 1363.06$\pm$423.0 & \textbf{1401.1$\pm$458.9}\\
										\midrule
										 &\multirow{2}{*}{\parbox[c]{1.5cm}{\centering div.-exp.-med.}} 
										& LowKnee & 1898.8$\pm$54.1 & 1689.5$\pm$304.2 & \textbf{1950.46$\pm$37.0} & \textbf{1892.8$\pm$161.7}\\
										& & LongHead & 1645.3$\pm$543.2 & 1498.2$\pm$727.3 & \textbf{1948.96$\pm$46.9} & 988.6$\pm$597.0 \\
										\midrule
										 &\multirow{2}{*}{\parbox[c]{1.5cm}{\centering div.-medium}} 
										& LowKnee & 1176.4$\pm$619.8 & \textbf{1797.9$\pm$166.7} & 1626.66$\pm$274.7 &  1359.1$\pm$380.0 \\
										& & LongHead & 1587.8$\pm$235.1 & 1737.9$\pm$88.7 & 1652.26$\pm$285.6 &\textbf{ 1859.5$\pm$154.0}\\
										\midrule
										 &\multirow{2}{*}{\parbox[c]{1.5cm}{\centering div.-med.-repl.}} 
										& LowKnee & 1593.2$\pm$439.9 & 1142.4$\pm$469.4 & \textbf{1780.16$\pm$336.1} & 1594.1$\pm$321.3\\
										& & LongHead & 1492.4$\pm$453.2 & 1146.1$\pm$584.8 & \textbf{1958.26$\pm$33.1} & \textbf{1941.8$\pm$62.2}\\
					\bottomrule
				\end{tabular}
			\end{sc}
		\end{small}
	\end{center}
	\vskip -0.2in
\end{table*}

\subsection{Few-shot Adaptation to Unseen Tasks}
\begin{figure}[tb]
	\centering
	\subfigure[HopperLowKnee.]{\includegraphics[width=0.2\columnwidth]{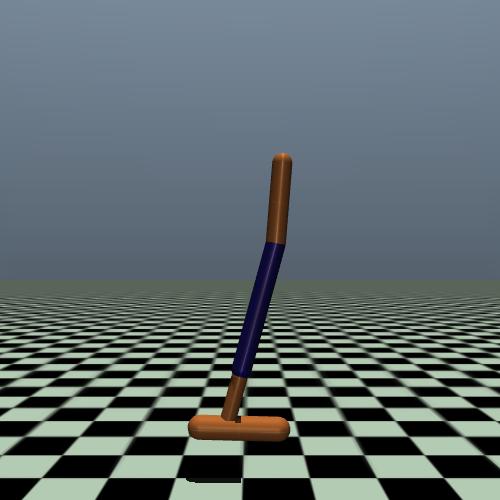}}
	\hfil
	\subfigure[HopperLongHead.]{\includegraphics[width=0.2\columnwidth]{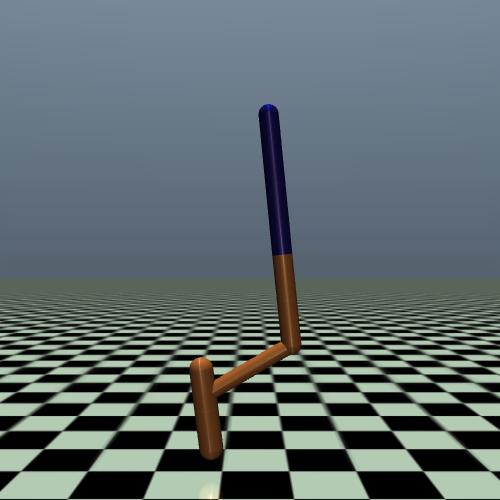}}
	\vspace{-0.3cm}
	\caption{Tasks for few-shot adaptation. Links modified from the original agent are indicated as blue. }
	\label{fig:fewshot-adaptation-tasks}
	\vspace{-0.3cm}
\end{figure}

The purpose of the few-shot adaptation experiment was to demonstrate that learning multiple behaviors from a single task in a training environment allows us to perform few-shot adaptation to a new environment by selecting a usable behavior from behaviors obtained in the training environment.
We prepared two environments for the few-shot adaptation experiment shown in Figure~\ref{fig:fewshot-adaptation-tasks}.
The embodiment of the agent, such as lengths of links, is changed on these tasks. 

In the few-shot adaptation experiment, policies trained in offline RL were adapted to a test environment different from the one used for training.
The protocol of the few-shot adaptation experiment was based on the one reported in \cite{Kumar20a}.
Given a latent-conditioned policy trained via offline RL, we sampled the value of the latent variable uniformly and executed the latent-conditioned policy in a test environment with the sampled value of the latent variable. 
In this phase, we tested each policy only once and evaluated the policy with $K$ different values of the latent variable.
In \cite{Kumar20a}, $K$ is called the budget for few-shot adaptation, and we set $K=25$ in our experiment.
The latent variable was continuous and two-dimensional in this experiment, and we uniformly sampled the value of the latent variable during the adaptation phase.
Additional details pertaining these tasks are provided in Appendix~\ref{app:fewshot_exp}.

The results of the few-shot adaptation experiments are summarized in Table~\ref{tbl:fewshot_results}.
The proposed methods outperformed baseline methods in this experiment.
These results correlate with the diversity scores shown in Table~\ref{tbl:diverse_results_diversity}, 
demonstrating that the proposed method successfully enables few-shot adaptation to new environments by learning diverse behaviors from a single task in offline RL.
These results also indicate that the proposed method discovers more diverse behaviors than the compared baseline methods.

\section{Conclusions}
In this study, we tackled the challenge of obtaining multiple solution in offline RL. 
We introduced a novel algorithm designed to learn diverse behaviors from individual tasks in offline RL, resembling a coordinate ascent approach akin to the EM algorithm. 
Experimental results illustrate that our proposed algorithm effectively learns multiple solutions in offline RL and that our method learn more diverse solutions than baseline methods.
Additionally, the findings suggest that the acquired multiple behaviors enables few-shot adaptation to new environments with limited test samples.
While we have presented a framework for formulating the problem of learning multiple solutions in offline RL, there is still room for improvement in enhancing the diversity of behaviors.
We hope to inspire future work to investigate alternative approaches to discover diverse solutions in offline RL.

\section*{Impact Statement}
This study was undertaken with the aim of contributing to the field of reinforcement learning. 
Reinforcement learning has various applications in society, and its impact is considered diverse. 
However, regarding this study, we believe there are no notable ethical issues or concerns.

\section*{Acknowledgements}
This work was partially supported by JST Moonshot R\&D Grant Number JPMJPS2011, CREST Grant Number JPMJCR2015, JSPS
KAKENHI Grant Number JP23K18476 and Basic Research Grant (Super AI) of Institute for AI and Beyond of the University of Tokyo.



\bibliography{diveoff}
\bibliographystyle{icml2024}

\newpage
\appendix
\onecolumn
\section{Derivation of the latent-conditioned policy update}
\label{app:derivation}
We can show the relationship between (7) and (8) in the main manuscript by extending the results presented in \cite{Peters10,Deisenroth13,Nair20}.
We consider an offline RL problem where a dataset is collected by a behavior policy that exhibits multiple behaviors.
Here, we assume that the behavior policy is given by 
\begin{align}
	\beta(\vect{a}|\vect{s}) = \int \beta(\vect{a}|\vect{s}, \vect{z})p(\vect{z})d\vect{z},
\end{align}
where $p(\vect{z})$ represents the prior distribution of the latent variable, and the value of $\vect{z}$ determines the behavior style of the behavior policy.
A fixed dataset can be regarded as samples generated from the distribution $d^\beta(\vect{s})\beta(\vect{a}|\vect{s})$, where $d^\beta(\vect{s})$ represents the stationary distribution of the state obtained by executing the behavior policy $\beta(\vect{a}|\vect{s})$.
For example, when a dataset is constructed by executing two distinctive policies, the latent variable is discrete, e.g. $z \in \{ 0, 1\}$, and the prior distribution $p(\vect{z})$ defines the ratio of executing these policies.

According to the above assumption, we consider the problem formulated as follows:
\begin{align}
	\max_{\pi} \E_{\vect{a}\sim\pi} \left[A_{\vect{w}}(\vect{s}, \vect{a}, \vect{z})\right] \\
	\textrm{s.t.} \ \ \  \KL \left( \pi(\vect{a}|\vect{s},\vect{z})||\beta(\vect{a}|\vect{s},\vect{z}) \right) \leq \epsilon \\
	\int \pi(\vect{a}|\vect{s}, \vect{z}) d\vect{a} = 1
	\label{eq:KL_RL_problem}
\end{align}
Here, we assume that the approximated posterior distribution $q(\vect{z}|\vect{s}, \vect{a})$ is given. 
The problem in  \eqref{eq:KL_RL_problem} is equivalent to (7) in the main manuscript when taking the expectation with respect to $d^\beta(\vect{s})$ and $q(\vect{z}|\vect{s})$.
The Lagrangian of the constraint problem in \eqref{eq:KL_RL_problem} is given by
\begin{align}
	\mathcal{L}(\pi) = \E_{\vect{a}\sim\pi} \left[A_{\vect{w}}(\vect{s}, \vect{a}, \vect{z})\right]  
	+ \alpha \left( \epsilon -  \KL \left( \pi(\vect{a}|\vect{s},\vect{z})||\beta(\vect{a}|\vect{s},\vect{z}) \right) \right) 
	+ \lambda \left( 1 -\int \pi(\vect{a}|\vect{s}, \vect{z}) d\vect{a} \right).
\end{align}
By taking the partial derivative of $\mathcal{L}(\pi)$ with respect to $\pi$, we obtain the following equation: 
\begin{align}
	\frac{\partial \mathcal{L}(\pi)}{ \partial \pi} =A_{\vect{w}}(\vect{s}, \vect{a}, \vect{z}) +  \alpha \left( - \log \beta(\vect{a}|\vect{s},\vect{z}) + \log \pi(\vect{a}|\vect{s},\vect{z}) + 1 \right)  - \lambda,
\end{align}
When we set $\frac{\partial \mathcal{L}(\pi)}{ \partial \pi}=0$, we obtain a non-parametric solution to the problem in \eqref{eq:KL_RL_problem}:
\begin{align}
	\pi^* \propto \beta(\vect{a}|\vect{s}, \vect{z}) \exp \left(\frac{1}{\alpha} A_{\vect{w}}(\vect{s}, \vect{a}, \vect{z})\right).
\end{align}
Given a dataset $\mathcal{D}=\{ (\vect{s}_i, \vect{a}_i) \}^N_{i=1} \sim d^\beta(\vect{s})\beta(\vect{a}|\vect{s})$ and the posterior distribution $q(\vect{z}|\vect{s}, \vect{a})$, we project the non-parametric solution to the parametric model $\pi_{\vect{\theta}}(\vect{a}|\vect{s}, \vect{z})$ by minimizing the KL divergence:
\begin{align}
	\min_{\vect{\theta}} \E_{\vect{s} \sim d^\beta, \vect{z} \sim q} \left[  \KL \left( \left. \beta(\vect{a}|\vect{s}, \vect{z})  \exp \left(\frac{1}{\alpha} A_{\vect{w}}(\vect{s}, \vect{a}, \vect{z})\right) \right\| \pi_{\vect{\theta}}(\vect{a}|\vect{s}, \vect{z})\right) \right].
	\label{eq:KL_latent_awr}
\end{align}
The KL divergence in \eqref{eq:KL_latent_awr} can be rewritten as
\begin{align}
	&\KL \left( \left. \beta(\vect{a}|\vect{s}, \vect{z})  \exp \left(\frac{1}{\alpha} A_{\vect{w}}(\vect{s}, \vect{a}, \vect{z})\right) \right\| \pi_{\vect{\theta}}(\vect{a}|\vect{s}, \vect{z})\right)  \nonumber \\
	& = \int \beta(\vect{a}|\vect{s}, \vect{z})  \exp \left(\frac{1}{\alpha} A_{\vect{w}}(\vect{s}, \vect{a}, \vect{z})\right) \log \beta(\vect{a}|\vect{s}, \vect{z})  \exp \left(\frac{1}{\alpha} A_{\vect{w}}(\vect{s}, \vect{a}, \vect{z})\right) d\vect{a} \nonumber \\
	& \ \ \ -  \int \beta(\vect{a}|\vect{s}, \vect{z})  \exp \left(\frac{1}{\alpha} A_{\vect{w}}(\vect{s}, \vect{a}, \vect{z})\right) \log \pi_{\vect{\theta}}(\vect{a}|\vect{s}, \vect{z}) d\vect{a}.
	\label{eq:KL_latent_awr2}
\end{align}
As the first term in the right-hand side of \eqref{eq:KL_latent_awr2} is independent of the policy $\pi_{\vect{\theta}}(\vect{a}|\vect{s}, \vect{z})$, the problem in \eqref{eq:KL_latent_awr} can be rewritten as
\begin{align}
	\max_{\vect{\theta}} \E_{\vect{s} \sim d^\beta, \vect{a}\sim\beta, \vect{z} \sim q} \left[\exp \left(\frac{1}{\alpha} A_{\vect{w}}(\vect{s}, \vect{a}, \vect{z})\right) \log \pi_{\vect{\theta}}(\vect{a}|\vect{s}, \vect{z})  \right].
\end{align}
Thus, we can obtain the relationship between (7) and (8) in the main manuscript.

%

\section{Details of diverse datasets}
\label{app:dataset}
\subsection{Details of tasks in diverse datasets}
In the reward function of the original tasks in D4RL, a velocity term $r_{\textrm{vel}}$ is used to encourage the agent to walk faster, and $r_{\textrm{vel}}$ is given by
\begin{align}
	r_{\textrm{vel}} = ( x_{t} - x_{t-1})) / \Delta t,
\end{align}
where $x_t$ represents the horizontal position of the agent at time $t$, and $\Delta t$ represents the time step size in the simulation.
To emphasize the existence of diverse solutions, we modified the velocity term of the reward function from that in the original tasks in the D4RL framework,
as in \cite{Kumar20}: 
\begin{align}
	r_{\textrm{vel}} = \min\big(  (x_{t} - x_{t-1}) / \Delta t, v_{\textrm{max}} \big),
\end{align}
where $x_t$ represents the horizontal position of the agent at time $t$, and $v_{\textrm{max}}$ is a constant term that defines the upper bound of $r_{\textrm{vel}}$.
The value of $v_{\textrm{max}}$ used in each agent is presented in Table~\ref{tbl:upper_bound_vel}.

The maximum return obtained using the trained policy and the minimum return obtained using a random policy are presented in Table~\ref{tbl:max_min_vel_tasks}.
\begin{table}[H]
	\caption{Upper bound for the velocity term in our dataset}
	\vskip 0.15in
	\label{tbl:upper_bound_vel}
	\begin{center}
		\begin{small}
			\begin{tabular}{lc}
				\hline
				Agent &  $v_{\textrm{max}}$  \\
				\hline
				Walker & 2\\
				Hopper & 1\\
				Halfcheetah & 2\\
				Ant & 1.5\\
				\hline
			\end{tabular}
		\end{small}
	\end{center}
\end{table}
\begin{table}[H]
	\caption{Maximum returns obtained using the trained policy and the minimum return obtained using a random policy}
	\vskip 0.15in
	\label{tbl:max_min_vel_tasks}
	\begin{center}
		\begin{small}
			\begin{tabular}{lcc}
				\hline
				Agent &  Max. return & Min. return  \\
				\hline
				Walker & 2860 & -4.02\\
				Hopper & 1962 & 6.84\\
				Halfcheetah & 1869 &-324.79\\
				Ant & 2265 & -379.33\\
				\hline
			\end{tabular}
		\end{small}
	\end{center}
\end{table}

\subsection{How to construct the diverse dataset}

We collected the samples by running LTD3 proposed in \cite{Osa22} according to the author implementation in \url{https://github.com/TakaOsa/LTD3}.
To create diverse datasets, we ran LTD3 five times with different random seeds with the two-dimensional continuous latent variable.
For ``diverse-expert'' and ``diverse-medium'' datasets, we collected samples by running the trained policies with uniformly generated values of the latent variable.
As the trained policy occasionally resulted in an episode with a low return, we removed such episodes from the ``diverse-expert'' and ``diverse-medium'' datasets.
The minimum return to be included in a dataset is presented in Table~\ref{tbl:threshold}.
The approximate number of samples in the ``diverse-expert'' and ``diverse-medium'' datasets was 300,000, although there was a variance in the number of samples due to the variance of the number of the steps in episodes.
The number of time steps for training policies for each dataset is presented in Table~\ref{tbl:steps}.
The ``medium-expert'' datasets were constructed by concatenating the ``expert'' and ``medium'' datasets.
Therefore, the approximate number of samples in the ``diverse-medium-expert'' datasets was 600,000.

To construct the ``medium-replay'' datasets, we aggregated the data in replay buffers from five different random seeds.
To avoid making the datasets too large, we did not include all the samples of the replay buffers in the ``medium-replay'' datasets .
We selected and included one episode after every $N$ episodes from the replay buffer, and the data from five random seeds were aggregated.
The value of $N$ was changed for each agent owing to the file size of the dataset.
The values of $N$ are presented in Table~\ref{tbl:replay_interval}.
The hyperparameters of LTD3 used to train policies for constructing our datasets are presented in Table~\ref{tbl:hyperparam_ltd3}.
We also provide visualization of the locomotion behaviors in the expert-diverse datasets in Figure~\ref{fig:diverse_expert}, which we omitted from the main manuscript owing to the page limitation.

\begin{table}[H]
	\caption{Minimum return for each dataset}
	\vskip 0.15in
	\label{tbl:threshold}
	\begin{center}
		\begin{small}
			\begin{tabular}{lcc}
				\hline
				Agent &  ``expert'' &  ``medium''   \\
				\hline
				Walker & 2000 & 500\\
				Hopper & 1700 & 500\\
				Halfcheetah & 1650 & 500\\
				Ant & 1700 & 500\\
				\hline
			\end{tabular}
		\end{small}
	\end{center}
\end{table}

\begin{table}[H]
	\caption{Training time steps to obtain policies for each dataset}
	\vskip 0.15in
	\label{tbl:steps}
	\begin{center}
		\begin{small}
			\begin{tabular}{lccc}
				\hline
				Agent &  ``expert'' &  ``medium'' & ``medium-replay   \\
				\hline
				Walker & 3 million & 1 million & 1 million\\
				Hopper & 3 million & 1 million & 1 million \\
				Halfcheetah & 2 million & 250,000 & 250,000\\
				Ant & 3 million & 1 million & 1 million\\
				\hline
			\end{tabular}
		\end{small}
	\end{center}
\end{table}

\begin{table}[H]
	\caption{Interval to be included in the medium-replay dataset}
	\vskip 0.05in
	\label{tbl:replay_interval}
	\begin{center}
		\begin{small}
			\begin{tabular}{lc}
				\hline
				Agent & $N$    \\
				\hline
				Walker & 5\\
				Hopper & 5\\
				Halfcheetah & 2 \\
				Ant & 5\\
				\hline
			\end{tabular}
		\end{small}
	\end{center}
		\vskip -0.1in
\end{table}

\subsection{Diversity of the samples in our datasets}
To quantify sample diversity within the datasets, we evaluated the entropy of their states. 
However, as the datasets' sample distributions are unknown, the entropy estimation is not a trivial task. 
Instead of directly estimating the entropy, we therefore evaluated its upper bound. 
When the density is approximated with a Gaussian mixture model as $p(\vect{s}) = \sum^M_{i=1} w_i \mathcal{N}(\vect{\mu}, \vect{\Sigma})$,
the upper bound of the entropy $\mathcal{H}(\vect{s})$ is given as follows:
\begin{align}
	\mathcal{H}(\vect{s})\leq \sum^L_{i=1} w_i \left( \frac{D}{2}\log (1 + 2\pi) + \frac{1}{2}\log \det (\vect{\Sigma}) - \log w_i \right)
	\label{eq:entropy_ub}
\end{align}

\begin{table}[H]
	\caption{ Diversity of datasets based on the state entropy.
		WK = Walker2d, HP = Hopper, HC = Halfcheetah.  }
	\label{tbl:dataset_diversity}
	\vskip 0.05in
	\begin{center}
		\begin{small}
			\begin{tabular}{l|rr|rr}
				\toprule
				& \parbox[c]{1.5cm}{\centering  diverse-expert-v0 \\ (ours)} & \parbox[c]{1.5cm}{\centering  expert-v2 \\ (d4rl)} & \parbox[c]{1.5cm}{\centering  diverse-medium-v0 (ours)} & \parbox[c]{1.5cm}{\centering  medium-v2 (d4rl)} \\
				\hline
				WK & \textbf{5.97$\pm$1.19} & -5.77$\pm$1.94 & \textbf{5.73$\pm$1.96} & 2.06$\pm$2.35 \\
				HP & \textbf{1.76$\pm$0.78} & -2.29$\pm$1.03 & \textbf{5.0$\pm$1.03} & -0.44$\pm$1.28 \\
				HC & \textbf{13.73$\pm$0.26} & 1.34$\pm$0.4  & \textbf{17.01$\pm$0.6} & 11.96$\pm$1.36 \\
				\bottomrule
				\toprule
				& \parbox[c]{1.5cm}{\centering  diverse-medium-expert-v0 (ours)} & \parbox[c]{1.5cm}{\centering  medium-expert-v2 (d4rl)} & \parbox[c]{1.5cm}{\centering  diverse-medium-replay-v0 (ours)} & \parbox[c]{1.5cm}{\centering  medium-replay-v2 (d4rl)} \\
				\hline
				WK &  \textbf{6.26$\pm$1.67} & 0.76$\pm$2.03 & 8.41$\pm$1.72 & 12.65$\pm$0.84 \\
				HP &  \textbf{4.35$\pm$0.68} & 2.25$\pm$0.68  & 6.31$\pm$1.17 & 6.88$\pm$0.67\\
				HC &  \textbf{14.41$\pm$0.19} & 7.86$\pm$1.34 & \textbf{18.93$\pm$0.47} & 14.11$\pm$0.44\\
				\bottomrule
			\end{tabular}
		\end{small}
	\end{center}
\end{table}

To compute the upper bound of $\mathcal{H}(\vect{s})$ in \eqref{eq:entropy_ub}, we selected a batch of samples from the dataset and fitted a Gaussian mixture model. 
The number of Gaussian components was determined based on the Bayesian information criteria.

We compared the upper bound of the entropy with that for existing D4RL datasets, with results listed in Tablee~\ref{tbl:dataset_diversity}.
As shown, our datasets exhibit higher values compared with existing D4RL datasets, indicating higher behavioral diversity.

\begin{figure}[H]
	\centering
	\subfigure{\includegraphics[width=\columnwidth]{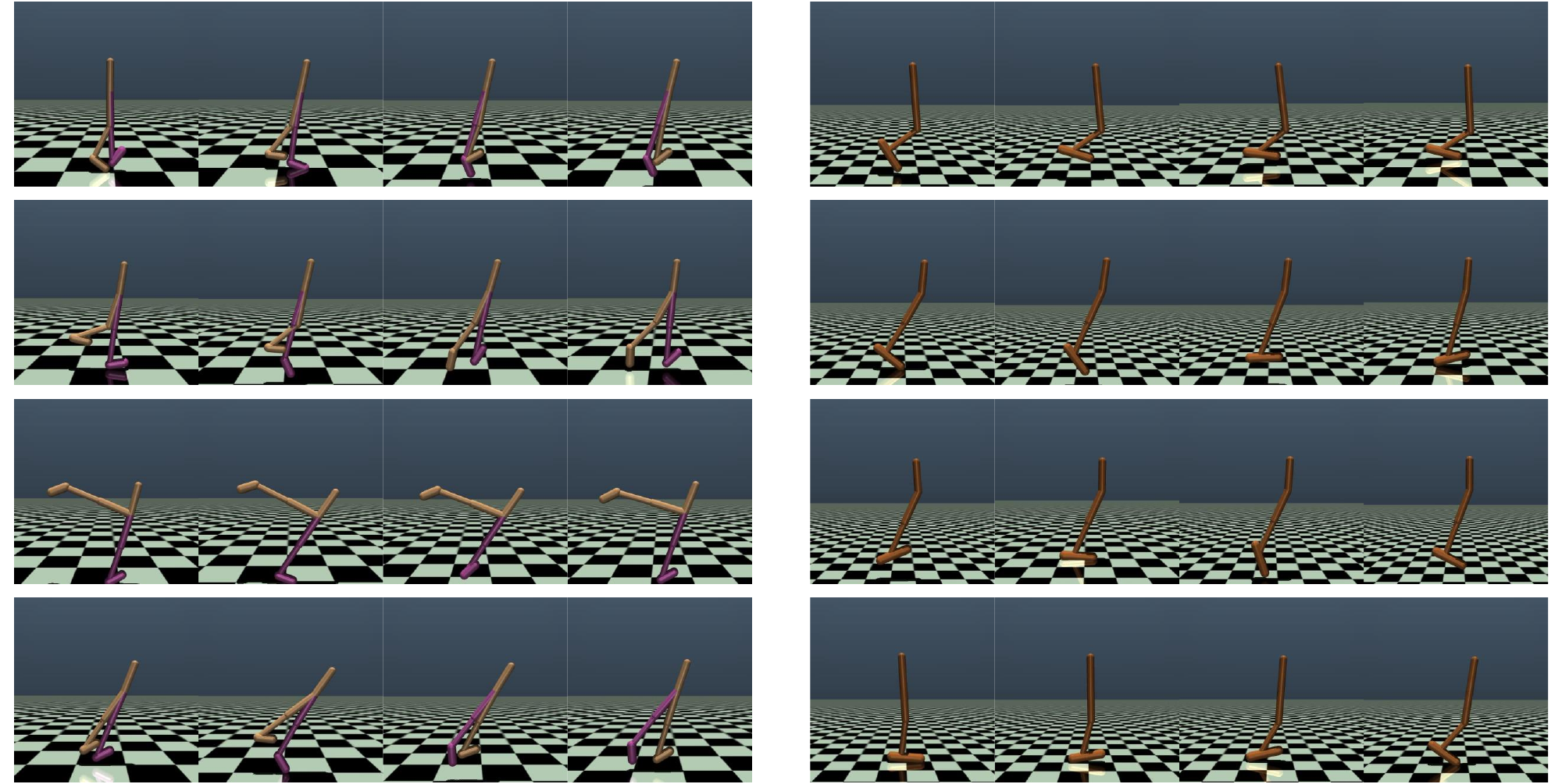}}
	\subfigure{\includegraphics[width=\columnwidth]{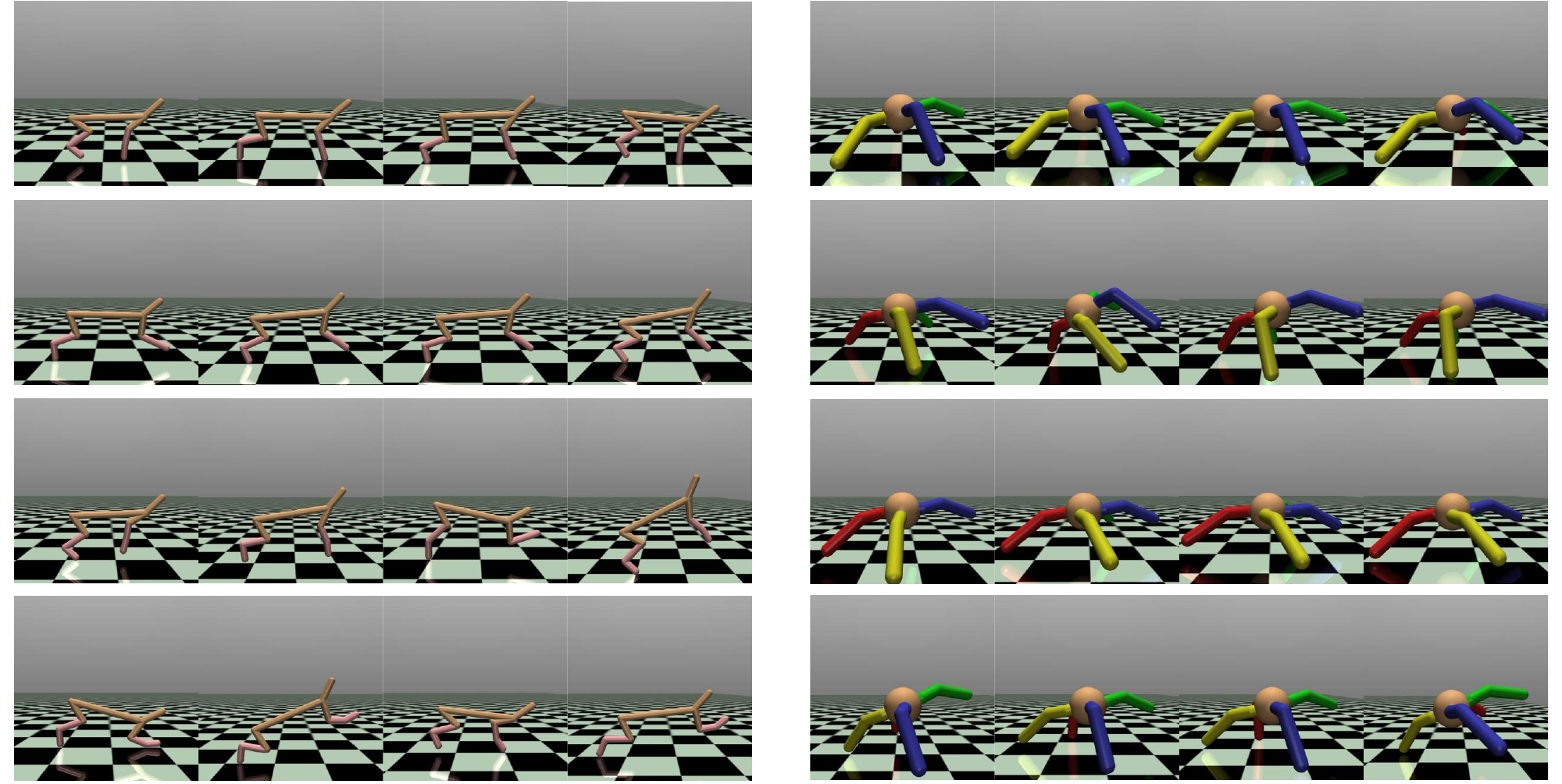}}
	\vspace{-0.7cm}
	\caption{Examples of locomotion behaviors contained in the expert-diverse datasets. The color of the legs of the ant agent was changed for visibility.  }
	\label{fig:diverse_expert}
	\vspace{-0.2cm}
\end{figure}

We also provide visualization of the locomotion behaviors in the expert-diverse datasets in Figure~\ref{fig:diverse_expert}, which we omitted from the main manuscript owing to the page limitation.

\begin{table}[H]
	\caption{Hyperparameters used for LTD3 to construct our datasets}
	\vskip 0.1in
	\label{tbl:hyperparam_ltd3}
	\begin{center}
		\begin{small}
			\begin{tabular}{lc}
				\hline
				Parameter & Value  \\
				\hline
				Optimizer    & Adam \\
				Learning rate & $3 \cdot 10^{-4}$ \\
				Discount factor $\gamma$ & 0.99 \\
				Replay buffer size & $10^6$ \\
				Number of hidden layers & 2 \\
				Number of hidden units per layer & 256 \\
				Number of samples per minibatch & 256 \\
				Activation function & Relu \\
				Target smoothing coefficient & 0.005 \\
				gradient steps per time step & 1 \\
				Clipping param. for IW & 0.3 \\
				\parbox[c]{5cm}{interval for maximizing  $\mathcal{J}_{\textrm{Q}}(\vect{\theta})$} & 2 \\
				\parbox[c]{5cm}{interval for maximizing $\mathcal{J}_{\textrm{info}}(\vect{\theta})$} & 5 for walker, 4 for others \\
				\hline
			\end{tabular}
		\end{small}
	\end{center}
\end{table}


\section{Diversity score}
\label{app:diversity_score}
To quantify the diversity of the learned solutions, we employed the diversity metric used in \cite{Parker-Holder20}.
Given a set of policies $\{ \pi_i \}^M_{i=1}$, the diversity metric is expressed as
\begin{align}
	D_{\textrm{div}} = \det \left( K(\vect{\phi}(\pi_i), \vect{\phi}(\pi_j))^M_{i,j=1} \right),
\end{align}
where $\vect{\phi}(\pi) \in \R^{l}$ is the behavior embedding of policy $\pi$, and $K: \R^l \times \R^l \mapsto \R $ is a kernel function.
Specifically, we used the squared-exponential kernel function given by 
\begin{align}
	k(\vect{\phi}(\pi_i), \vect{\phi}(\pi_j)) = \exp \left( - \frac{ \left\| \vect{\phi}(\pi_i) - \vect{\phi}(\pi_j) \right\|^2 }{2h^2} \right),
\end{align}
where $\vect{\phi}(\pi_i)$ is the policy embedding given by the mean of the states visited by the policies as $\vect{\phi}(\pi_i) = \E_{\vect{s} \sim \pi_i, \mathcal{P}}\left[ \vect{s} \right]$.
We uniformly sampled the value of the latent variable and computed the policy embedding, with $M=9$. 
In the main manuscript, we reported the average and the standard deviation over five different random seeds.

\section{Comparison with methods that learn a single solution}
\label{app:comp_single}
We evaluated the performance of prevalent offline RL methods that finds a single solution on our datasets. 
Specifically, we considered behavior cloning, TD3+BC~\cite{Fujimoto21}, IQL~\cite{Kostrikov22}, AWAC~\cite{Nair20}, and LAPO~\cite{Chen22} as baseline methods that learn a single behavior. 
For TD3+BC, we used the author-provided implementations.
In our implementation of AWAC, we used the state normalization and double-clipped Q-learning as in TD3+BC.
We implemented IQL by adopting the author-provided implementation of IQL to pytorch.
We report the performance of policies after training with 1 million steps unless otherwise stated.
For DiveOff, a learning rate for $\mathcal{L}_{\textrm{M-info}}$ was $6e-4$ for all the task, and the D4RL score is different from Table~\ref{tbl:diverse_results_score}.

In Table~\ref{tbl:diverse_results_app}, DiveOff (ave.) is the average over different behaviors, while DiveOff (best) represents the results when we report the results of the latent variables that led to the highest return. 
To select the latent variable for DiveOff (best), we uniformly sampled the value of the latent variable 25 times, and the value that led to the highest total returns in three test episodes was used as the best value.

As shown, DiveOff (best) outperforms DiveOff (ave), and DiveOff (best) achieved a result comparable or superior to prevalent offline RL methods that learn a single policy.
Although learning multiple behaviors potentially increase the computational and algorithmic complexity, 
the performance of the proposed method in terms of the normalized score was found to be comparable to that of the baseline methods.
LAPO is designed to deal with multimodal distributions in a given dataset, and outperformed other baselines in diverse-expert datasets.
Notably, DiveOff achieved a performance comparable to LAPO on diverse-expert datasets, and the overall performance of DiveOff was better than LAPO.

Interestingly, the performance of IQL on the walker2d-diverse-expert and hopper-diverse-expert datasets was lower than that of other baseline methods, 
whereas IQL often achieved state-of-the-art performance on the expert dataset in D4RL.
However, IQL significantly outperformed other baseline methods on diverse-medium-replay datasets, indicating the ``stitching'' ability of IQL.

\begin{table*}[t]
	\caption{ Results on diverse dataset.
		WK = Walker2dVel, HP = HopperVel, HC = HalfcheetahVel, AN = AntVel. Best results are denoted in bold. The average and standard deviation of normalized scores over the 10 test episodes and five seeds are shown. }
	\label{tbl:diverse_results_app}
	\vskip 0.15in
	\begin{center}
		\begin{small}
			\begin{sc}
				\begin{tabular}{clrrrrr|rr}
					\toprule
					& & BC & TD3+BC & IQL & AWAC & LAPO & \parbox[c]{1.5cm}{\centering  DiveOff (ave.) } & \parbox[c]{1.5cm}{\centering  DiveOff (best) }  \\
					\midrule
					\multirow{4}{*}{\rotatebox[origin=c]{90}{\parbox[c]{1.5cm}{\centering  div.-exp.}}} 
					& WK & 95.0$\pm$4.8 & 75.5$\pm$31.9 & 23.5$\pm$31.0 & 90.8$\pm$3.3 & 93.1$\pm$7.4  & 98.5$\pm$2.2 & \textbf{99.8$\pm$0.5} \\
					& HP & \textbf{99.2$\pm$0.8} & 94.6$\pm$11.5 & 66.5$\pm$39.4 & 92.5$\pm$14.1 & \textbf{100.6$\pm$0.1}  & 96.4$\pm$3.8 & 99.7$\pm$0.6  \\
					& HC & 95.9$\pm$0.7 & 97.6$\pm$0.2 & \textbf{98.4$\pm$0.5} & 96.5$\pm$0.2 & 97.5$\pm$0.2 & 96.8$\pm$0.1 & 96.7$\pm$0.3  \\ 
					& AN& 93.6$\pm$2.6 & \textbf{98.5$\pm$2.8} & 97.1$\pm$2.1 & 93.8$\pm$2.4 & 96.6$\pm$1.0 & 95.4$\pm$1.2 & 95.9$\pm$0.6 \\
					\midrule
					\multirow{4}{*}{\rotatebox[origin=c]{90}{\parbox[c]{1.5cm}{\centering  div.-med.-exp.}}} 
					& WK & 63.5$\pm$24.8 & 96.5$\pm$7.0 & 49.1$\pm$36.5 & 74.7$\pm$22.2& 68.5$\pm$21.5 & \textbf{98.0$\pm$2.5} & \textbf{98.8$\pm$1.8} \\
					& HP & 98.1$\pm$4.9 & 92.1$\pm$14.0 & \textbf{101.1$\pm$0.0} & 79.1$\pm$23.9 & 100.2$\pm$0.5  & 98.9$\pm$2.2 & 98.9$\pm$1.7 \\
					& HC & 95.8$\pm$0.8 & \textbf{97.3$\pm$0.2} & \textbf{97.1$\pm$4.2} & 95.9$\pm$0.6 & \textbf{97.4$\pm$0.2} & 96.2$\pm$0.1 & 96.4$\pm$0.2 \\
					& AN& 79.5$\pm$6.7 & \textbf{96.2$\pm$3.8} &  91.3$\pm$5.2 & 85.1$\pm$3.7 & 94.8$\pm$1.2 & 88.5$\pm$1.9 & 88.9$\pm$1.1\\
					\midrule
					\multirow{4}{*}{\rotatebox[origin=c]{90}{\parbox[c]{1.5cm}{\centering  div.-med.}}} 
					& WK & 76.3$\pm$2.6 & 77.0$\pm$13.7 & 80.1$\pm$8.1 & 76.9$\pm$11.3 & \textbf{90.3$\pm$4.8} & \textbf{88.4$\pm$9.3} & \textbf{89.1$\pm$5.6} \\
					& HP & 73.5$\pm$24.4 & 88.1$\pm$9.0 & 94.5$\pm$6.2 & \textbf{96.9$\pm$3.0} & \textbf{98.2$\pm$3.4}  & 95.1$\pm$6.2 & 95.6$\pm$5.3\\
					& HC & 92.6$\pm$0.2 & 92.3$\pm$2.3 & \textbf{94.5$\pm$1.8}& 92.9$\pm$0.4&\textbf{94.4$\pm$0.2} & 93.2$\pm$0.1 & 93.3$\pm$0.3 \\
					& AN& 75.6$\pm$4.3 & 64.3$\pm$64.1 & \textbf{91.9$\pm$0.3} & 79.9$\pm$3.4 & 83.3$\pm$3.7 & 81.1$\pm$2.3 & 81.8$\pm$1.0 \\
					\midrule
					\multirow{4}{*}{\rotatebox[origin=c]{90}{\parbox[c]{1.5cm}{\centering  div.-med.-rep.}}} 
					& WK & 26.0$\pm$22.9 & 80.6$\pm$28.4 & \textbf{83.4$\pm$20.1} & 32.8$\pm$15.5 & 46.7$\pm$18.0 & 44.2$\pm$18.2 & 52.5$\pm$23.0 \\
					& HP & 26.5$\pm$5.2 & 67.9$\pm$23.1 & \textbf{101.1$\pm$0.0} & 100.9$\pm$0.2 & 77.0$\pm$6.1  & \textbf{101.1$\pm$0.1} & \textbf{101.1$\pm$0.1}\\
					& HC & 14.6$\pm$0.2 & \textbf{93.2$\pm$0.6} &\textbf{92.8$\pm$0.4} & 86.5$\pm$7.3 & 44.6$\pm$24.8 & 85.6$\pm$11.0 & 87.0$\pm$7.5 \\
					& AN& 32.9$\pm$1.7 & \textbf{91.9$\pm$3.2} & 91.4$\pm$0.6 &  35.0$\pm$2.5 & 37.6$\pm$6.4 & 32.9$\pm$2.9 & 36.0$\pm$2.5  \\
					\bottomrule
				\end{tabular}
			\end{sc}
		\end{small}
	\end{center}
	\vskip -0.1in
\end{table*}

\section{Comparison of variants of the baseline and proposed method}
\label{app:diveoff_iql}
As additional baseline, we evaluated the policy update based on TD3+BC, which we refer to as TD3-L+BC+VAE.
In TD3-L+BC+VAE, we used the latent-conditioned behavioral cloning term, and the objective function for TD3-L+BC+VAE is given by
\begin{align}
	\mathcal{L}_{\textrm{TD3+VAE}}(\vect{\theta}) = \E_{\vect{s} \sim \mathcal{D}} \E_{\vect{z}\sim q} \left[ Q_{\vect{w}}(\vect{s}, \vect{\mu}_{\vect{\theta}}(\vect{s}, \vect{z}), \vect{z})\right] \nonumber \\
	 - \lambda \E_{(\vect{s}, \vect{a}) \sim \mathcal{D}} \E_{\vect{z}\sim q} \left[ \left( \vect{\mu}_{\vect{\theta}}(\vect{s}, \vect{z}) - \vect{a} \right)^2 \right], 
	 \label{eq:td3-vae}
\end{align}
where $\lambda$ is a constant, and $q$ is the posterior distribution as a part of VAE.
To implement TD3-L+BC+VAE, we mofieid the author-provided implementation of TD3+BC.
We evaluated TD3-L+BC+VAE with $\lambda \in {1.0, 2.5, 5.0, 10.0 }$ in \eqref{eq:td3-vae}, but did not observe significant differences. 
Therefore, we report the results with $\lambda=2.5$ as in the default setting in TD3+BC.

While we reported the performance of DiveOff based on the double-clipped Q-learning in the main text, alternative methods can be employed.
To investigate the effect of the algorithm for learning the critic, we evaluate a variant of DiveOff using the quantile regression as in IQL~\cite{Kostrikov22}.

The results are reported in Tables~\ref{tbl:quantile_score} and \ref{tbl:quantile_diversity}.

\begin{table}[tb]
	\vskip -0.15in
		\caption{ Normalized scores of methods that train latent-conditioned policies.
			WK = Walker2dVel, HP = HopperVel, HC = HalfcheetahVel, AN=AntVel. Best results are denoted in bold. 
		}
		\label{tbl:quantile_score}
		\vskip -0.15in
		\begin{center}
			\begin{small}
				\begin{sc}
					\begin{tabular}{clrrr}
						\toprule
						& & \parbox[r]{2.cm}{ TD3+BC+VAE } & \parbox[r]{2.cm}{ DiveOff (quantile regress.)} & \parbox[c]{2.cm}{ DiveOff (double clipped Q) }   \\
						\midrule
						\multirow{4}{*}{\rotatebox[origin=c]{90}{\parbox[c]{1.5cm}{\centering  div.-exp.}}} 
						& WK & 3.7$\pm$3.6 & 27.7$\pm$35.9 & \textbf{93.3$\pm$9.9}\\
						& HP & 0.7$\pm$0.2 & 6.6$\pm$8.4 & \textbf{81.2$\pm$33.8}\\
						& HC & 14.6$\pm$0.1 & 94.2$\pm$8.7 & \textbf{96.8$\pm$0.3}\\
						& AN & 51.5$\pm$0.4 &\textbf{99.1$\pm$0.3} & 95.4$\pm$1.2 \\
						\midrule
						\multirow{4}{*}{\rotatebox[origin=c]{90}{\parbox[c]{1.5cm}{\centering  div.-exp.-med.}}} 
						& WK & 4.7$\pm$3.5 & 49.4$\pm$30.2 & \textbf{80.5$\pm$17.7} \\
						& HP & 0.8$\pm$0.2 & \textbf{101.1$\pm$0.0} & 98.4$\pm$4.7 \\
						& HC & 14.6$\pm$0.0 & \textbf{97.4$\pm$4.2} &  96.1$\pm$0.6\\
						& AN & 51.7$\pm$0.3 & \textbf{96.9$\pm$0.4} & 88.5$\pm$1.9 \\
						\midrule
						\multirow{4}{*}{\rotatebox[origin=c]{90}{\parbox[c]{1.5cm}{\centering  div.-med.}}} 
						& WK & 4.1$\pm$3.3 & 62.8$\pm$20.9 &  \textbf{67.9$\pm$17.4}\\
						& HP & 0.8$\pm$0.3 & \textbf{99.7$\pm$2.7} & 83.1$\pm$18.0 \\
						& HC & 14.6$\pm$0.0 & \textbf{95.5$\pm$0.4} &  93.2$\pm$0.5\\
						& AN & 51.7$\pm$0.2 & \textbf{92.9$\pm$0.8} & 81.1$\pm$2.3 \\
						\midrule
						\multirow{4}{*}{\rotatebox[origin=c]{90}{\parbox[c]{1.5cm}{\centering  div.-med.-rep.}}} 
						& WK & 3.4$\pm$3.8 & \textbf{90.8$\pm$7.0} &  39.3$\pm$11.6\\
						& HP & 1.1$\pm$0.4 & \textbf{101.1$\pm$0.0} &  \textbf{101.0$\pm$0.1}\\
						& HC & 14.7$\pm$0.0 & \textbf{93.5$\pm$0.5} &  89.8$\pm$6.4\\
						& AN & 51.8$\pm$0.3 & \textbf{95.0$\pm$0.7} &  32.9$\pm$2.9\\
						\bottomrule
					\end{tabular}
				\end{sc}
			\end{small}
		\end{center}
		\vskip -0.1in
	\end{table}

\begin{table}[tb]
	\caption{ Diversity scores of methods that train latent-conditioned policies.
		WK = Walker2dVel, HP = HopperVel, HC = HalfcheetahVel, AN=AntVel. Best results are denoted in bold. The results for cases where the D4RL scores fall below 20.0 compared to the best score are presented in gray text. The results for cases where the D4RL scores are the best among the compared methods are presented with underlines.
	}
	\label{tbl:quantile_diversity}
	\vskip -0.15in
	\begin{center}
		\begin{small}
			\begin{sc}
					\begin{tabular}{clrrr}
					\toprule
					& & \parbox[c]{2.cm}{\centering TD3+BC + VAE } & \parbox[c]{2.cm}{\centering DiveOff (quantile regress.)} & \parbox[c]{2.cm}{\centering DiveOff (double clipped Q) }   \\
					\midrule
					\multirow{4}{*}{\rotatebox[origin=c]{90}{\parbox[c]{1.5cm}{\centering  div.-exp.}}} 
					& WK & \gray{0.24$\pm$0.12} & 0.025$\pm$0.05 & \underline{0.22$\pm$0.23}\\
					& HP & \gray{0.71$\pm$0.36} & 0.07$\pm$0.10 & \underline{\textbf{0.88$\pm$0.11}}\\
					& HC & \gray{0.09$\pm$0.08} & \textbf{0.73$\pm$0.28} & \underline{0.77$\pm$0.24}\\
					& AN & \gray{0.05$\pm$0.08} & \underline{\textbf{0.14$\pm$0.18}} & 0.09$\pm$0.11 \\
					\midrule
					\multirow{4}{*}{\rotatebox[origin=c]{90}{\parbox[c]{1.5cm}{\centering  div.-exp.-med.}}} 
					& WK & \gray{0.25$\pm$0.21} & 0.46$\pm$0.20 & \underline{\textbf{0.75$\pm$0.20}} \\
					& HP & \gray{0.65$\pm$0.37} & \underline{0.05$\pm$0.09} &  \textbf{0.88$\pm$00.19}\\
					& HC & \gray{0.06$\pm$0.06} & \underline{0.07$\pm$0.05} & \textbf{0.96$\pm$0.04} \\
					& AN & \gray{0.28$\pm$0.30} & \underline{0.06$\pm$0.06} & \textbf{0.61$\pm$0.27} \\
					\midrule
					\multirow{4}{*}{\rotatebox[origin=c]{90}{\parbox[c]{1.5cm}{\centering  div.-med.}}} 
					& WK & \gray{0.32$\pm$0.30} & \textbf{0.43$\pm$0.35} & \underline{0.36$\pm$0.33} \\
					& HP & \gray{ 0.66$\pm$0.34} & \underline{0.68$\pm$0.29} & \textbf{0.79$\pm$0.34} \\
					& HC & \gray{0.04$\pm$0.04} & \underline{0.12$\pm$0.08} & \underline{0.81$\pm$0.13} \\
					& AN & \gray{0.23$\pm$0.29} & \underline{0.09$\pm$0.12} & \textbf{0.65$\pm$0.26} \\
					\midrule
					\multirow{4}{*}{\rotatebox[origin=c]{90}{\parbox[c]{1.5cm}{\centering  div.-med.-rep.}}} 
					& WK & \gray{0.21$\pm$0.39} & \textbf{0.11$\pm$0.14} & \gray{0.01$\pm$0.01} \\
					& HP & \gray{0.66$\pm$0.37} & 0.001$\pm$0.001 &0.02$\pm$0.03 \\
					& HC & \gray{0.004$\pm$0.005} & \textbf{0.29$\pm$0.35} & \gray{0.00$\pm$0.00} \\
					& AN & \gray{0.07$\pm$0.12} & 0.0$\pm$0.0 & \gray{0.06$\pm$0.12} \\
					\bottomrule
				\end{tabular}
			\end{sc}
		\end{small}
	\end{center}
	\vskip -0.15in
	\end{table}
	
The results indicate that replacing double-clipped-Q-learning with IQL improves the performance on the replay dataset. However, the performance on some expert datasets got worse. This effect can also be observed by comparing AWAC-L+VAE and IQL-L+VAE in Table~\ref{tbl:diverse_results_score}. Previous studies indicate that IQL has better stitching capability compared to double-clipped-Q-learning. The performance gain on the replay datasets implies that the stitching capability is necessary to learn from the replay datasets.

\section{Details of few-shot adaptation experiments}
\label{app:fewshot_exp}
\subsection{Environments}
The purpose of the few-shot adaptation experiment was to demonstrate that learning multiple behaviors from a single task in a training environment allows us to perform few-shot adaptation to a new environment by selecting a usable behavior from behaviors obtained in the training environment.
We prepared two environments for the few-shot adaptation experiments.
For the hopper agent, we prepared the \texttt{HopperLowKnee-v0} and \texttt{HopperLongHead-v0} environments.
In HopperLowKnee-v0, the position of the knee joint was lower than that of the normal hopper agent.
In HopperLongHead-v0, the top link was longer than that of the normal hopper agent.

\subsection{Procedure of the few-shot adaptation experiment}
In the few-shot adaptation experiment, policies trained in offline RL were adapted to a test environment different from the one used for training.
The protocol of the few-shot adaptation experiment was based on the one reported in \cite{Kumar20a}.
Given a latent-conditioned policy trained via offline RL, we sampled the value of the latent variable uniformly and executed the latent-conditioned policy in a test environment with the sampled value of the latent variable. 
In this phase, we tested each policy only once and evaluated the policy with $K$ different values of the latent variable.
In \cite{Kumar20a}, $K$ is called the budget for few-shot adaptation, and we set $K=25$ in our experiment.
Then, we stored the value of the latent variable $\vect{z}_{\textrm{max}}$ that resulted in the maximum return.
Subsequently, we ran the latent-conditioned policy with $\vect{z} = \vect{z}_{\textrm{max}}$ in the test environment 10 times and recorded the average return.
We repeated the above process five times for different random seeds, and the corresponding average and standard deviation are presented in Table~\ref{tbl:fewshot_results}.

\section{Implementation details and hyperparameters}
In DiveOff and AWAC, we used the state normalization employed in TD3+BC~\cite{Fujimoto21} and the double-clipped Q-learning~\cite{Fujimoto18}.
The neural network architecture and hyperparameters are presented in Table~\ref{tbl:diveoff_param}.

\begin{table}[tb]
	\caption{Hyperparameters of DiveOff. }
	\label{tbl:diveoff_param}
	\begin{center}
		\begin{small}
			\begin{tabular}{ll}
				\toprule
				Hyperparameter & Value \\
				\midrule
				Optimizer & Adam \\
				Learning rate for critic & 3e-4 \\
				Learning rate for actor $\pi_{\vect{\theta}}$ & 3e-4 \\
				Learning rate for posterior distribution $q_{\vect{\phi}}$ & 3e-4 (initialization) \\
				Learning rate for Likelihood $p_{\vect{\psi}}$ & 3e-4 (initialization) \\
				\parbox[c]{4.0cm}{ Learning rate for $\mathcal{L}_{\textrm{M-info}}$ } & 6e-5 for antvel, 9e-5 for others \\
				Mini-batch size & 256 \\
				Discount factor & 0.99 \\
				Target update rate & 5e-3 \\
				Policy update frequency & 2 \\
				Score scaling $1/\alpha_\pi$ & 3.0 \\
				Score scaling $1/\alpha_q$ & 1.0 \\
				\midrule
				Critic activation function & ReLU \\
				Actor activation function & ReLU \\
				Encoder hidden dim & 256 \\
				Encoder hidden layers & 2 \\
				Encoder activation function & ReLU \\
				Decoder hidden dim & 256 \\
				Decoder hidden layers & 2 \\
				Decoder activation function & ReLU \\
				\midrule
				$\lambda$ & 2.0 \\
				\bottomrule
			\end{tabular}
		\end{small}
	\end{center}
\end{table}

\begin{table}[tb]
	\caption{Hyperparameters of TD3-L+BC+VAE. }
	\label{tbl:td3_bc_param}
	\begin{center}
		\begin{small}
			\begin{tabular}{ll}
				\toprule
				Hyperparameter & Value \\
				\midrule
				Optimizer & Adam \\
				Critic learning rate & 3e-4 \\
				Actor learning rate & 3e-4 \\
				Mini-batch size & 256 \\
				Discount factor & 0.99 \\
				Target update rate & 5e-3 \\
				Policy noise & 0.2 \\
				Policy noise clipping & (-0.5, 0.5) \\
				Policy update frequency & 2 \\
				\midrule
				Critic hidden dim & 256 \\
				Critic hidden layers & 2 \\
				Critic activation function & ReLU \\
				Actor hidden dim & 256 \\
				Actor hidden layers & 2 \\
				Actor activation function & ReLU \\
				Encoder hidden dim & 256 \\
				Encoder hidden layers & 2 \\
				Encoder activation function & ReLU \\
				Decoder hidden dim & 256 \\
				Decoder hidden layers & 2 \\
				Decoder activation function & ReLU \\
				\midrule
				$\alpha$ & 2.5 \\
				\bottomrule
			\end{tabular}
		\end{small}
	\end{center}
\end{table}

\begin{table}[tb]
	\caption{Hyperparameters of AWAC-L+VAE. }
	\label{tbl:awac_param}
	\begin{center}
		\begin{small}
			\begin{tabular}{ll}
				\toprule
				AWAC Hyperparameter & Value \\
				\midrule
				Optimizer & Adam \\
				Critic learning rate & 3e-4 \\
				Actor learning rate & 3e-4 \\
				Mini-batch size & 256 \\
				Discount factor & 0.99 \\
				Target update rate & 5e-3 \\
				Policy update frequency & 2 \\
				Score scaling $1/\alpha$ & 3.0 \\
				\midrule
				Critic hidden dim & 256 \\
				Critic hidden layers & 2 \\
				Critic activation function & ReLU \\
				Actor hidden dim & 256 \\
				Actor hidden layers & 2 \\
				Actor activation function & ReLU \\
				Encoder hidden dim & 256 \\
				Encoder hidden layers & 2 \\
				Encoder activation function & ReLU \\
				Decoder hidden dim & 256 \\
				Decoder hidden layers & 2 \\
				Decoder activation function & ReLU \\
				\bottomrule
			\end{tabular}
		\end{small}
	\end{center}
\end{table}

\begin{table}[tb]
	\caption{Hyperparameters of IQL-L+VAE. }
	\label{tbl:IQL_param}
	\begin{center}
		\begin{small}
			\begin{tabular}{ll}
				\toprule
				IQL Hyperparameter & Value \\
				\midrule
				Optimizer & Adam \\
				Critic learning rate & 3e-4 \\
				Actor learning rate & 3e-4 \\
				Mini-batch size & 256 \\
				Discount factor & 0.99 \\
				Target update rate & 5e-3 \\
				Score scaling $1/\alpha$ & 3.0 \\
				Expectile $\tau$ & 0.7 \\
				\midrule
				Critic hidden dim & 256 \\
				Critic hidden layers & 2 \\
				Critic activation function & ReLU \\
				Actor hidden dim & 256 \\
				Actor hidden layers & 2 \\
				Actor activation function & ReLU \\
				Encoder hidden dim & 256 \\
				Encoder hidden layers & 2 \\
				Encoder activation function & ReLU \\
				Decoder hidden dim & 256 \\
				Decoder hidden layers & 2 \\
				Decoder activation function & ReLU \\
				\bottomrule
			\end{tabular}
		\end{small}
	\end{center}
\end{table}

\section{Effect of hyperarameters}
We provide the effect of the learning rate of the optimizer for $\mathcal{L}_{\textrm{M-info}}$ in Table~\ref{tbl:hyper_diversity_app}.
The rest of the hyperparamters were the same as in Table~\ref{tbl:diveoff_param}.
As we fixed the learning rate for other objective functions, changing the learning rate for $\mathcal{L}_{\textrm{M-info}}$ corresponds to the change of the weight on $\mathcal{L}_{\textrm{M-info}}$.
The higher learning rate leads to the larger weight on $\mathcal{L}_{\textrm{M-info}}$.
As shown in Table~\ref{tbl:hyper_diversity_app}, we observed the trade-off between the D4RL score and diversity score.

\begin{table*}[t]
	\caption{ Effect of the learning rate of $\mathcal{L}_{\textrm{M-info}}$.
		WK = Walker2dVel, HP = HopperVel, HC = HalfcheetahVel. AN = AntVel. Best results are denoted in bold. The average and standard deviation of normalized scores over the 10 test episodes and five seeds are shown. }
	\label{tbl:hyper_diversity_app}
	\vskip 0.15in
	\begin{center}
		\begin{small}
			\begin{sc}
				\begin{tabular}{clrr}
					\toprule
					\multicolumn{4}{c}{D4RL score} \\
					\midrule
					& & \parbox[c]{1.5cm}{\centering  DiveOff lr=9e-5 } & \parbox[c]{1.5cm}{\centering  DiveOff lr=6e-5 }  \\
					\midrule
					\multirow{4}{*}{\rotatebox[origin=c]{90}{\parbox[c]{1.5cm}{\centering  div.-exp.}}} 
					& WK & 93.3$\pm$9.9 & \textbf{98.5$\pm$2.2} \\
					& HP & 81.2$\pm$33.8 &\textbf{96.4$\pm$3.8}   \\
					& HC & 96.8$\pm$0.3 &96.8$\pm$0.1   \\ 
					& AN & 95.6$\pm$0.6 &95.4$\pm$1.2  \\
					\midrule
					\multirow{4}{*}{\rotatebox[origin=c]{90}{\parbox[c]{1.5cm}{\centering  div.-med.-exp.}}} 
					& WK & 80.5$\pm$17.7 &\textbf{98.0$\pm$2.5}  \\
					& HP & 98.4$\pm$4.7 &98.9$\pm$2.2  \\
					& HC & 96.1$\pm$0.6 &96.2$\pm$0.1 \\
					& AN & 83.6$\pm$3.7 &88.5$\pm$1.9 \\
					\midrule
					\multirow{4}{*}{\rotatebox[origin=c]{90}{\parbox[c]{1.5cm}{\centering  div.-med.}}} 
					& WK & 67.9$\pm$17.4 &\textbf{88.4$\pm$9.3}  \\
					& HP & 83.1$\pm$18.0 &\textbf{95.1$\pm$6.2} \\
					& HC & 93.2$\pm$0.5 &93.2$\pm$0.1  \\
					& AN & 76.7$\pm$8.9 &\textbf{81.1$\pm$2.3}  \\
					\midrule
					\multirow{4}{*}{\rotatebox[origin=c]{90}{\parbox[c]{1.5cm}{\centering  div.-med.-rep.}}} 
					& WK & 39.3$\pm$11.6 &44.2$\pm$18.2  \\
					& HP & 101.0$\pm$0.1 &101.1$\pm$0.1 \\
					& HC & 89.8$\pm$6.4 &85.6$\pm$11.0  \\
					& AN & 33.9$\pm$2.4 &32.9$\pm$2.9   \\
					\bottomrule
				\end{tabular}
				\hfil
				\begin{tabular}{clrr}
					\toprule
					\multicolumn{4}{c}{Diversity score} \\
					\midrule
					& & \parbox[c]{1.5cm}{\centering  DiveOff lr=9e-5 } & \parbox[c]{1.5cm}{\centering  DiveOff lr=6e-5 }  \\
					\midrule
					\multirow{4}{*}{\rotatebox[origin=c]{90}{\parbox[c]{1.5cm}{\centering  div.-exp.}}} 
					& WK & \textbf{0.220$\pm$0.230} & 0.146$\pm$0.169\\
					& HP & \textbf{0.879$\pm$0.114} & 0.441$\pm$0.331  \\
					& HC & 0.774$\pm$0.241 & 0.868$\pm$0.208  \\ 
					& AN & \textbf{0.148$\pm$0.142} & 0.090$\pm$0.111 \\
					\midrule
					\multirow{4}{*}{\rotatebox[origin=c]{90}{\parbox[c]{1.5cm}{\centering  div.-med.-exp.}}} 
					& WK & \textbf{0.751$\pm$0.195} & 0.132$\pm$0.140 \\
					& HP & \textbf{0.878$\pm$0.189} &  0.115$\pm$0.130\\
					& HC & \textbf{0.961$\pm$0.041} &  0.742$\pm$0.297\\
					& AN & 0.323$\pm$0.317 & \textbf{0.612$\pm$0.267}\\
					\midrule
					\multirow{4}{*}{\rotatebox[origin=c]{90}{\parbox[c]{1.5cm}{\centering  div.-med.}}} 
					& WK & \textbf{0.364$\pm$0.327} & 0.022$\pm$0.019 \\
					& HP & 0.788$\pm$0.338 & 0.707$\pm$0.327\\
					& HC & \textbf{0.813$\pm$0.131} &  0.460$\pm$0.187\\
					& AN & 0.278$\pm$0.221 & \textbf{0.645$\pm$0.259} \\
					\midrule
					\multirow{4}{*}{\rotatebox[origin=c]{90}{\parbox[c]{1.5cm}{\centering  div.-med.-rep.}}} 
					& WK & 0.007$\pm$0.009 & \textbf{0.083$\pm$0.166} \\
					& HP & \textbf{0.015$\pm$0.030} & 0.000$\pm$0.000\\
					& HC & 0.003$\pm$0.003 & 0.003$\pm$0.005 \\
					& AN & \textbf{0.244$\pm$0.386} & 0.062$\pm$0.120  \\
					\bottomrule
				\end{tabular}
			\end{sc}
		\end{small}
	\end{center}
	\vskip -0.1in
\end{table*}


\end{document}